\documentclass[10pt,twocolumn,letterpaper]{article}

\usepackage[T1]{fontenc}
\usepackage[utf8]{inputenc}

\usepackage{iccv}
\usepackage{times}
\usepackage{epsfig}
\usepackage{graphicx}
\usepackage{amsmath}
\usepackage{amssymb}
\usepackage{amsthm}

\usepackage[usenames,dvipsnames]{xcolor}  
\usepackage{tcolorbox}
\usepackage{enumitem,amssymb,amsmath,pifont}
\usepackage{booktabs}
\usepackage{tabularx}
\usepackage{multirow}
\usepackage{adjustbox}
\usepackage{subfig}
\usepackage{wrapfig}
\usepackage{afterpage}
\usepackage{mdframed}
\usepackage{arydshln} 
\usepackage{array}
\newcolumntype{H}{>{\setbox0=\hbox\bgroup}c<{\egroup}@{}}

\newcommand{\vcenteredinclude}[1]{
    \includegraphics[height=2.5\fontcharht\font`\B]{#1}
}

\newcommand{\emoji}{%
    \begingroup\normalfont
    \hspace*{-1\fontcharht\font`\B}
    \raisebox{-1.2ex}{\vcenteredinclude{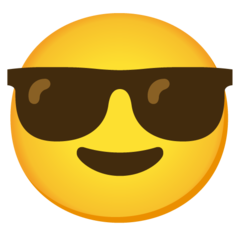}}
    \hspace*{-1.5\fontcharht\font`\B}
    \endgroup
}

\newcommand{\pietruszkacolor}[1]{\textcolor{black}{#1}}

\newcommand{\jordyskeleton}[1]{\textcolor{black}{#1}}

\newlist{todolist}{itemize}{2}
\setlist[todolist]{label=$\square$}
\usepackage{todonotes}
\usepackage{float}
\usepackage{color, colortbl}
\usepackage{etoc}
\usepackage{pdfpages}

\newcommand*{\avgpmstd}[2]{%
{#1}\raisebox{0.15ex}{\scriptsize±{#2}}
}

\usepackage[pagebackref=true,breaklinks=true,letterpaper=true,colorlinks,bookmarks=false]{hyperref}
\usepackage{cleveref}

\iccvfinalcopy 

\def\iccvPaperID{12222}

\ificcvfinal\fi

\newcommand{\projectwithemoji}{\textbf{DUDE}~\emoji{}~} 
\newcommand{\project}{\textbf{DUDE}} 
\newcommand{\ECE}{$\mathrm{ECE}$} 
\newcommand{\AURC}{$\mathrm{AURC}$} 
\newcommand{\ANLS}{$\mathrm{ANLS}$} 

\begin{document}
\definecolor{Gray}{rgb}{0.9, 0.9, 0.9}

\title{Document Understanding Dataset and Evaluation\\(\projectwithemoji{})}

\newcommand{\superaffil}[2]{\textsuperscript{#1}\,#2}

\author{
\small Jordy Van Landeghem\superaffil{1,2}\and
\small Rubèn Tito\superaffil{5}\and
\small Łukasz Borchmann\superaffil{3}\and
\small Michał Pietruszka\superaffil{3,6}\and
\small Paweł Józiak\superaffil{3,4}\and
\small Rafał Powalski\superaffil{8}\and
\small Dawid Jurkiewicz\superaffil{3,7}\and
\small Mickaël Coustaty\superaffil{9}\and
\small Bertrand Ackaert\superaffil{2}\and
\small Ernest Valveny\superaffil{5}\and
\small Matthew Blaschko\superaffil{1}\and
\small Sien Moens\superaffil{1}\and
\small Tomasz Stanisławek\superaffil{3}
\and
\footnotesize{ 
  \textsuperscript{1}KU Leuven
  \quad
  \textsuperscript{2}Contract.fit
  \quad
  \textsuperscript{3}Snowflake
  \quad
  \textsuperscript{4}Warsaw University of Technology
  \quad
  \textsuperscript{5}Computer Vision Center, Universitat Autònoma de Barcelona}\\
\footnotesize{
  \quad
  \textsuperscript{6}Jagiellonian University
  \quad
  \textsuperscript{7}Adam Mickiewicz University
  \quad
  \textsuperscript{8}Instabase
  \quad
  \textsuperscript{9}University of La Rochelle
  }
}

\maketitle

\ificcvfinal\thispagestyle{empty}\fi

\etocdepthtag.toc{mtchapter}
\etocsettagdepth{mtchapter}{subsection}
\etocsettagdepth{mtappendix}{none}
\begin{abstract}
We call on the Document AI (DocAI) community to re-evaluate current methodologies and embrace the challenge of creating more practically-oriented benchmarks. Document Understanding Dataset and Evaluation (\project{}) seeks to remediate the halted research progress in understanding visually-rich documents (VRDs). We present a new dataset\footnote{\url{huggingface.co/datasets/jordyvl/DUDE_loader}} with novelties related to types of questions, answers, and document layouts based on \textbf{multi-industry}, \textbf{multi-domain}, and \textbf{multi-page} VRDs of various origins, and dates.
Moreover, we are pushing the boundaries of current methods by creating multi-task and multi-domain evaluation setups that more accurately simulate real-world situations where powerful generalization and adaptation under low-resource settings are desired.
\project{} aims to set a new standard as a more practical, long-standing benchmark for the community, and we hope that it will lead to future extensions and contributions that address real-world challenges. Finally, our work illustrates the importance of finding more efficient ways to model language, images, and layout in DocAI.

\end{abstract}

\section{Introduction} 
Early stages of research and growth in any field are characterized by enacting proof-of-concept and demonstrating the feasibility of the proposed solution. In the Deep Learning era, this is often echoed by building narrow and simplified datasets that do not reflect real-world complexity, leading to models that may not be suitable for practical use. 

The field of Document Understanding (DU) is not an exception to the recent proliferation of deep architectures, which in this case are predominantly used for classification and information extraction from documents. However, the wide and complex nature of documents presents many challenges that remain unsolved or not yet addressed. One such challenge is domain generalization, where a model trained on medical documents may not be directly applicable to financial or tabular content. Another challenge concerns task-agnostic ar\-chi\-tec\-tu\-res, where a model must be able to adapt to various DU subtasks such as document classification, key information extraction (KIE), and question answering (QA). Lastly, the high variability of document contents and layouts often leads to highly imbalanced samples within document types, resulting in a long-tailed distribution with few or almost no samples to train a model.

Despite the importance of these challenges, there is currently no DU benchmark dataset that simultaneously addresses all of these issues.
This paper proposes a novel dataset formulated as an instance of Document Visual Question Answering (DocVQA) to evaluate how well current DU solutions deal with multi-page documents, if they can navigate and reason over visual layouts, and if they can generalize their skills to different document types and domains.

The data collection and evaluation design of \project{} naturally motivates targeting models that can answer natural yet highly diverse questions (e.g., regarding document elements, their properties, and compositions) for any VRD (e.g., drawn from potentially unseen distributions of layouts, domains, and types). The presented problem setting relates to Multi-Domain Long-Tailed Recognition (MDLT) \cite{yang2022multi}, which concerns learning from multi-domain imbalanced data whilst addressing label imbalance, divergent label distributions across domains, and possible train-test domain shift. Put plainly, since we cannot provide ground truth QA pairs for, e.g., stamps, on every document type (domain), we expect a solution to transfer the subtask 'stamp detection' learned on document types where stamps naturally occur (and thus training QA pairs were created organically) to other domains.
The DocVQA and MDLT formulations of \project{} allow us to create a longstanding, challenging benchmark that in the future can be easily extended with more subtasks formulated as QA pairs, and domains relating to document types (see Limitations).

The contribution of this work is twofold. First, we have created \project{}, a novel large-scale, multi-paged, multi-domain, multi-industry DocVQA benchmark for evaluating DU progress. Second, we show that the zero-shot and fine-tuned performance of current state-of-the-art models applied to DU lags far behind human baselines, explained in part by the need for more holistic and efficient modeling of language, vision, and richly structured layouts.

\section{Related Work}
Document Understanding encompasses datasets related to various subtasks like document layout analysis~\cite{zhong2019publaynet, li2020docbank}, classification~\cite{harley2015evaluation}, key information extraction~\cite{kleisterStanislawekGWLK21,jaume2019funsd}, table extraction~\cite{smock2022pubtables,zhong2020image,zheng2021global}, and visual question answering~\cite{mathew2020document, mishra2019ocr, tito2021icdar}.
These benchmarks lead to end-to-end DU architectures that have transformed common DocAI practices \cite{Powalski2021GoingFB,appalaraju2021docformer,huang2022layoutlmv3, Garncarek2021LAMBERTLL,gu2021unidoc,li2021selfdoc,https://doi.org/10.48550/arxiv.2206.04045}. These task-specific benchmarks, however, are often tailored to a single domain, limiting the ability to create and assess how well DU models generalize to other document types and domains. To fill this gap, we adopt a visual question answering (VQA) approach, which has been crucial in the growth of the DU field. 

\looseness=-1 The VQA paradigm provides a natural language interface for various tasks from both computer vision and natural language processing. In the latter, the question-answering approach has been successfully used in several domains, including medicine~\cite{pappas-etal-2020-biomrc,kamath:hal-01759306,Nentidis_2022,jin-etal-2019-pubmedqa,li-etal-2021-mlec,raghavan-etal-2021-emrkbqa,moller-etal-2020-covid}, open-domain knowledge~\cite{yang-etal-2015-wikiqa,https://doi.org/10.48550/arxiv.2007.15207,https://doi.org/10.48550/arxiv.2004.10645,liu-etal-2019-xqa}, emotions~\cite{gui-etal-2017-question,bjerva-etal-2020-subjqa}, code~\cite{agashe-etal-2019-juice,liu-wan-2021-codeqa-question}, logical reasoning~\cite{ijcai2020p0501,yu2020reclor,zhang2022nail,https://doi.org/10.48550/arxiv.2204.07408}, claim verification~\cite{thorne-etal-2018-fever,hu-etal-2022-chef,zarharan-etal-2021-parsfever}, and math~\cite{zhang-etal-2021-noahqa-numerical,hopkins-etal-2019-semeval,chen-etal-2021-geoqa,mishra-etal-2022-numglue,https://doi.org/10.48550/arxiv.1905.13319}. As a result of its ability to function as a natural language interface for various forms of data, this paradigm has been applied to other domains. For example, the question-answering approach is combined with modalities such as videos~\cite{lei-etal-2018-tvqa, castro-etal-2020-lifeqa, castro-etal-2022-wild, gupta-demner-fushman-2022-overview, colas-etal-2020-tutorialvqa}, images~\cite{yoshikawa-etal-2017-stair, https://doi.org/10.48550/arxiv.1505.00468,https://doi.org/10.48550/arxiv.1802.08218, https://doi.org/10.48550/arxiv.2004.10796,biten2019icdar, biten2019scene}, speech~\cite{you-etal-2022-end, lakomkin-etal-2018-kt}, knowledge graphs~\cite{trivedi2017lc,10.1145/3340531.3412760, saxena-etal-2021-question, dutt-etal-2022-perkgqa, kacupaj-etal-2021-conversational}, and maps~\cite{paz-argaman-tsarfaty-2019-run,https://doi.org/10.48550/arxiv.2211.08545}.

\looseness=-1 Overall, the convergence of computer vision and NLP through the emergence of VQA tasks has also opened up new avenues for research in the DU field, with many DU datasets now including rich visual content alongside questions.
Yet, prior study on document VQA has mainly focused on single-page documents~\cite{mathew2020document, tito2021document, mathew2022infographicvqa} with rare exceptions such as MP-DocVQA~\cite{tito2022hierarchical}. 
However, ~\cite{mathew2020document,tito2021document} pose only extractive questions where the answer follows the context on which the question is defined as in other question answering benchmarks~\cite{rajpurkar2016squad, trischler2016newsqa,kwiatkowski2019natural}. Moreover, these datasets do not contain \textit{non-answerable} questions as in established (natural language) QA datasets like ~\cite{rajpurkar2018know,kwiatkowski2019natural}. To the best of our knowledge, there are no VQA datasets containing questions requiring lists as an answer. There are however few text-only QA datasets that contain such answer types~\cite{pasupat2015compositional,li2022multispanqa,dasigi2019quoref}.
Other datasets mainly related to our work are rather domain-specific like ~\cite{zhu2022towards, VisualMRC2021, mathew2022infographicvqa, SlideVQA, qi-etal-2022-dureadervis}. We give a detailed comparison of most related document VQA datasets in \Cref{tab:dataset_summary} highlighting the major contributions.





\section{\project{} Dataset}

While \project{} shares some similarities with existing VQA datasets, a closer comparison (see Table~\ref{tab:dataset_summary}) highlights its unique features. We are confident that the model's proficiency in the areas introduced in this work will showcase its capability to handle the intricacy and diversity of document understanding tasks in real-world scenarios.

\begin{figure}
    \centering
    \includegraphics[width=0.85\linewidth,trim={2.5cm 1.5cm 0.8cm 1.6cm},clip]{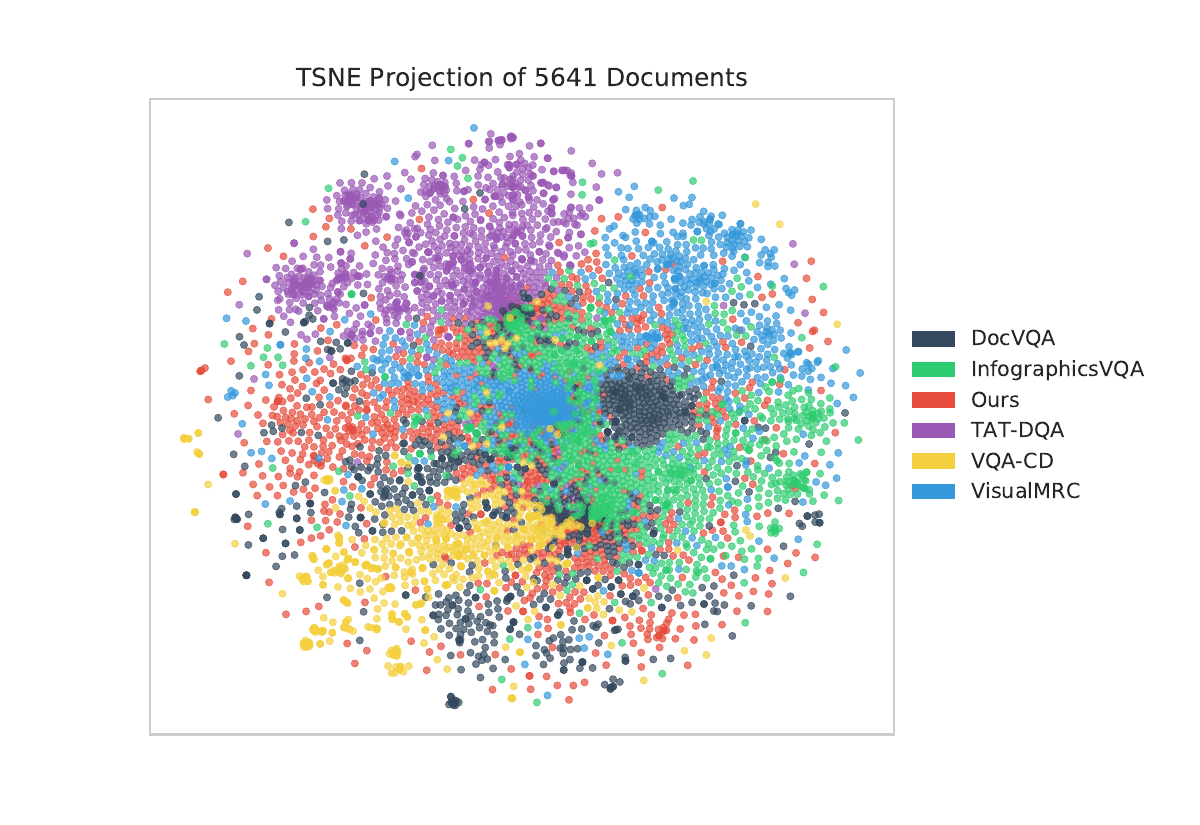}
    \caption{Visualization of inter-document similarities between samples from different datasets (t-SNE over TF-IDF representations of 1k passages from each source).}
    \label{fig:vocab}
\end{figure}

\paragraph{Documents.} 
The dataset covers a wide range of document types, sources and dates, as shown in \Cref{tab:dataset_summary} and \Cref{fig:vocab} where its diverse nature is confirmed by the spread of document content representations.\footnote{This holds not only when textual content is considered but also for document images (\Cref{fig:layout} in the Appendix).} Moreover, it covers a broad range of domains, including medical, legal, technical, and financial, among others, to evaluate models' ability to handle diverse topics and the specific knowledge each requires. Furthermore, the dataset contains documents with varying layouts: diverse text arrangements, font sizes, and styles, to ensure that models can handle visually diverse documents.

In contrast to our proposal, current VQA datasets often focus on homogeneous documents, such as invoices in VQA-CD~\cite{10.1007/978-3-031-06555-2_44} or financial reports in TAT-DQA~\cite{zhu2022towards}. Even when not restricted to a single domain or layout, these datasets share essential characteristics. For example, InfographicsVQA~\cite{mathew2022infographicvqa} demonstrates significant diversity in topics and designs, but still embodies a preference for visual aids over complex tables or long text passages.
Moreover, VQA datasets are commonly restricted to either born-digital or scanned documents, which limits their ability to measure the robustness to mixed-origin files that one usually finds in real-world applications. 
In particular, this restriction makes it uncertain whether state-of-the-art performers on website fragments from VisualMRC \cite{VisualMRC2021} can be efficient on multi-column layouts and documents with OCR errors or incorrectly-detected reading orders.
Finally, a typical dataset for document visual question answering contains documents from a limited period, i.e., a few years (Table~\ref{tab:dataset_summary}).

Considering the properties mentioned above,  the most diverse dataset to date is Single Page DocVQA (SP-DocVQA) \cite{mathew2020document}, which contains mixed-origin documents of different types created over several decades. However, it is built exclusively on single-page document excerpts and is limited to several domains represented in the Industry Documents Library. As a result, it complements rather than serves as a touchstone for general-purpose DU systems. MP-DocVQA~\cite{tito2022hierarchical} extends this including previous and posterior pages of the documents. However, the questions are kept the same which makes the extra pages mere distractors.

\paragraph{Questions.}
We use VQA as a natural language interface to VRDs, challenging the DU model with diverse questions, advanced operations, and multi-step reasoning to achieve real-world success.

Firstly, we assert that various layouts and visual elements must be comprehended semantically. As such, we introduce complex questions targeting these document elements, 
requiring comprehension beyond the document content, such as `\textsl{how many text columns are there?}', `\textsl{does the document contain words with diacritics}?' or `\textsl{which page contains the largest table in the document?}'. These Layout-navigating questions bridge the gap between Document Layout Analysis and Question Answering paradigms.

Our unique and detailed compositional questions demand a model that comprehends semantics and generalizes to new questions in a zero-shot setting.
For example, >90\% of our questions are unique, while we target questions whose answer scope is much more diverse than in previous works.\footnote{Answer type comparison is included in supplementary materials.}
Since neural networks are known to perform poorly at mathematical reasoning and symbolical processing, we provide training and evaluation questions demanding arithmetic and comparison operations on numbers and dates.

Moreover, we feature multi-hop questions that indicate a model's robustness to sequential reasoning and mimic how humans ask questions. They may be useful in real-world tasks such as `\textsl{If the checkbox on page 1 section 3a indicates that the company is incorporated, how much yearly revenue did it generate in 2022 (given the table on page 5)}?'

\paragraph{Answers.} Even though some VQA datasets are deliberately limited to questions of exclusively extractive (SP-DocVQA) or abstractive (VisualMRC) nature, others do not obey such restrictions and include both question types (see Table~\ref{tab:dataset_summary}). The dataset we provide includes both abstractive and extractive answers, covering various types such as \textit{textual, numerical, dates, yes/no, lists, or no answer}.

This allows us to cover all possible business use cases and reveal major deficiencies of existing DU systems beyond typical textual answers. For instance, no existing VQA dataset includes not answerable questions and questions answered with a list. In turn, the models considered to date supposedly tend to make unreliable guesses on questions with an answer not entailed by the content \cite{rajpurkar2018know}. Our dataset is designed to cover answers beyond plain extractive text such as a list of items or even `None'.

The `None' answer type demands that the model correctly identifies that the answer cannot be provided, as the question needs to be better formed, e.g., it asks about the value of an empty cell in the table. In addition, list generation problems pose challenges to the model, as (1) more tokens need to be generated, (2) they may be sourced from different places in the document, and (3) OCR reading order may influence the element ordering.

\subsection{Gathering Documents}\label{sec:gathering_documents}

A fundamental difficulty in gathering raw source files was ensuring dataset diversity while fulfilling strict licensing requirements. Therefore, rather than depending on initial sources of files, e.g., libraries that originally published digitized materials, we resorted to aggregate websites. 

The document collection process was manual and assumed formulating queries to \href{http://archive.org}{archive.org} (containing 36M books and texts), \href{http://commons.wikimedia.org}{commons.wikimedia.org} (with 86M media types of various types), and \href{http://documentcloud.org}{documentcloud.org} (with around 5M public documents). 
The queries consisted of keywords relevant to some category of interest, e.g., the \textit{resume} category of our proposal consists of `resume', `cv', `curriculum', and `biography' keywords). Where necessary, a separate query parameter ensured that the resulting files belonged to the public domain or were released under a permissive license. Information on keywords and the search procedure is distributed as a part of the DUDE dataset. 

From the resulting documents, we selected those representing the requested category and visually distinctive from the ones already gathered. Special care was put into removing examples that visibly expose controversial content or may be subject to privacy or legal concerns, despite the declared license. We collected five thousand, typically multi-page, English documents using this methodology.

\subsection{Annotation Process}
The annotation process involved in-house annotators and Amazon Mechanical Turk freelancers. For the latter, there is limited control over the expertise, and where justified, we resorted to limiting task availability depending on the number of completed tasks and historical acceptance rate.\footnote{Approval above 97\% over at least 5k HITs.} The former are five highly qualified people with a Ph.D. in Linguistics. These three annotation scenarios will be referred to as \textit{All MTurkers}, \textit{Best MTurkers}, and \textit{Qualified Linguists}.

We estimate the total cost of annotation involving both \textit{Linguists} and \textit{MTurkers} as \$20,000.

\paragraph{Phase 1.} \looseness=-1  We started by providing \textit{All MTurkers} documents described in Section~\ref{sec:gathering_documents} in separate batches aimed at collecting abstractive, extractive, and list QA pairs. Each freelancer was asked to propose up to five questions of a particular type, and in the case of extractive ones to provide an evidence bounding box. The exception to this process is the annotation of non-answerable questions previously shown to be particularly challenging \cite{rajpurkar2018know}. These are predominantly annotated by \textit{Qualified Linguists} and because of their quality promoted without passing through Phases 2-3.

\looseness=-1 Candidate QA pairs are semi-automatically filtered to exclude annotations that cannot be valid due to the length, use of non-typical character combinations, or type-specific criteria, such as non-list answers for list batches. Additionally, we cluster duplicate and near-duplicate question-answer pairs to ensure dataset diversity and promote them directly to Phase~3 after a manual review (the same QA pairs provided independently by several annotators indicate their validity).

\paragraph{Phase 2.} The rest of the annotations promoted from Phase~1 were directed to \textit{All MTurkers}, but this time instead of providing complete QA pairs, they were asked to answer the question from the previous round. Obtained triples of questions and two answer variants (one from each phase) were evaluated using inter-answer $\mathrm{ANLS}$ (defined in \Cref{sec:evaluation}) promoted to the final dataset if the agreement was >0.8. Otherwise, QA triples were directed to Phase 3. 

\paragraph{Phase 3.} \textit{Best MTurkers} were provided with document, question, and answer variants to decide the correctness of each answer and optionally overrule both variants if they are not correct. Outliers from decisions in this phase, such as repealing without a judgment on previous answers, were reviewed by \textit{Qualified Linguists} and corrected if needed.

\paragraph{Optional Phase 4.} Annotations of the test set were reviewed by \textit{Qualified Linguists}. Given data from Phase~3, they corrected questions, answers and created metadata related to diagnostic categories described in Section~\ref{sec:diagnostic}.


\subsection{Dataset Statistics} 

\begin{table*}[ht]
\begin{adjustbox}{width=\textwidth}
\small
\bgroup
\def\arraystretch{0.9}
\begin{tabular}{p{3.2cm}p{2.2cm}p{2.2cm}p{2.2cm}p{2.3cm}p{2.1cm}}
\toprule
Dataset & Ours & SP-DocVQA & VisualMRC & InfographicsVQA & TAT-DQA \\
\midrule
\multicolumn{6}{c}{\textit{Dataset-level properties}} \vspace{0.1cm}\\
Sources & Multi & Industry docs & Web pages & Infographics & Finance reports \\
Origin & BD, Scan & Mostly scans & BD & BD & BD \\
Period & 1860-2022 & 1960-2000 & Jan-Mar 2020 & not specified & 2018-2020 \\
Documents & 5,019 & 12,767 & 10,234 & 5,485 & 2,758 \\
Pages (\emph{\avgpmstd{avg}{std}}) & \avgpmstd{5.72}{6.4} & \avgpmstd{1.0}{0.0} & \avgpmstd{1.0}{0.0} & \avgpmstd{1.0}{0.0} & \avgpmstd{1.11}{0.32} \\
Tokens (\emph{\avgpmstd{avg}{std}}) & \avgpmstd{1,831.53}{2,545.06} & \avgpmstd{183}{149.96} & \avgpmstd{154.19}{79.34} & \avgpmstd{287.98}{214.57} & \avgpmstd{576.99}{290.12} \\
Simpson coeff. (ResNet) & {0.82} & {0.76} & {0.83} & {0.86} & {0.73} \\
Simpson coeff. (Tf-Idf) & {0.95} & {0.93} & {0.99} & {0.94} & {0.15} \\
\midrule
\multicolumn{6}{c}{\textit{Question-level properties}} \vspace{0.1cm}\\
Questions & 41,541 & 50,000 & 30,562 & 30,035 & 16,558 \\
Unique (\%) & 90.9 & 72.34 & 96.26 & 99.11 & 95.65 \\
Length (\emph{\avgpmstd{avg}{std}}) & \avgpmstd{8.65}{3.35} & \avgpmstd{8.34}{3.04} & \avgpmstd{9.38}{4.01} & \avgpmstd{11.57}{3.71} & \avgpmstd{12.51}{4.18} \\
Semantics & All & T, L, F, Ch & T, L, F, Ch & T, L, F, Ch, M & T, L \\ 
\midrule
\multicolumn{6}{c}{\textit{Answer-level properties}} \vspace{0.1cm}\\
Unique (\%) & 70.7 & 64.29 & 91.82 & 48.84 & 77.54 \\
Length (\emph{\avgpmstd{avg}{std}}) & \avgpmstd{3.35}{6.1} & \avgpmstd{2.11}{1.67} & \avgpmstd{8.38}{6.36} & \avgpmstd{1.66}{1.43} & \avgpmstd{3.44}{7.20} \\
Extractive (\%) & 42.39 & 100.0 & 0.0 & 71.96 & 55.72 \\
Abstractive (\%) & 38.25 & 0.0 & 100.0 & 24.91 & 44.28 \\
List (\%) & 6.62 & 0.0 & 0.0 & 5.69 & 0.0 \\
None & 12.74 & 0.0 & 0.0 & 0.0 & 0.0 \\
\bottomrule\\
\end{tabular}
\egroup
\end{adjustbox}

\caption{Summary of the existing English document datasets and our challenge. \texttt{BD} stands for born-digital. Layout semantics are abbreviated as (T)able, (L)ist, (F)igure, (Ch)art, and M(ap). Comparison based on Azure Cognitive Services (3.2) OCR. 
\label{tab:dataset_summary}}

\end{table*}

\looseness=-1 We conducted a statistical analysis of our dataset and found that the distribution of document length, question length, and answer type was much more diverse than in other datasets in the same domain. We also used the Simpson diversity coefficient~\cite{SIMPSON_1949} for analysis and summarized the results in Table~\ref{tab:dataset_summary}. The following are the statistics for the data split:

\begin{table}[H]
    \small
    \centering
    \begin{tabular}{c|ccc}
    \toprule
        {} & train & val & test (diagnostic) \\
        \midrule
        documents & 3,010 & 749 & 1,215 (530)\\
        questions & 23,728 & 6,315 & 11,448 (2,462)\\
        \bottomrule
    \end{tabular}
    \caption{Data split counts.}
\end{table}
\begin{figure*}[ht]
\centering
\includegraphics[width=.4\linewidth]{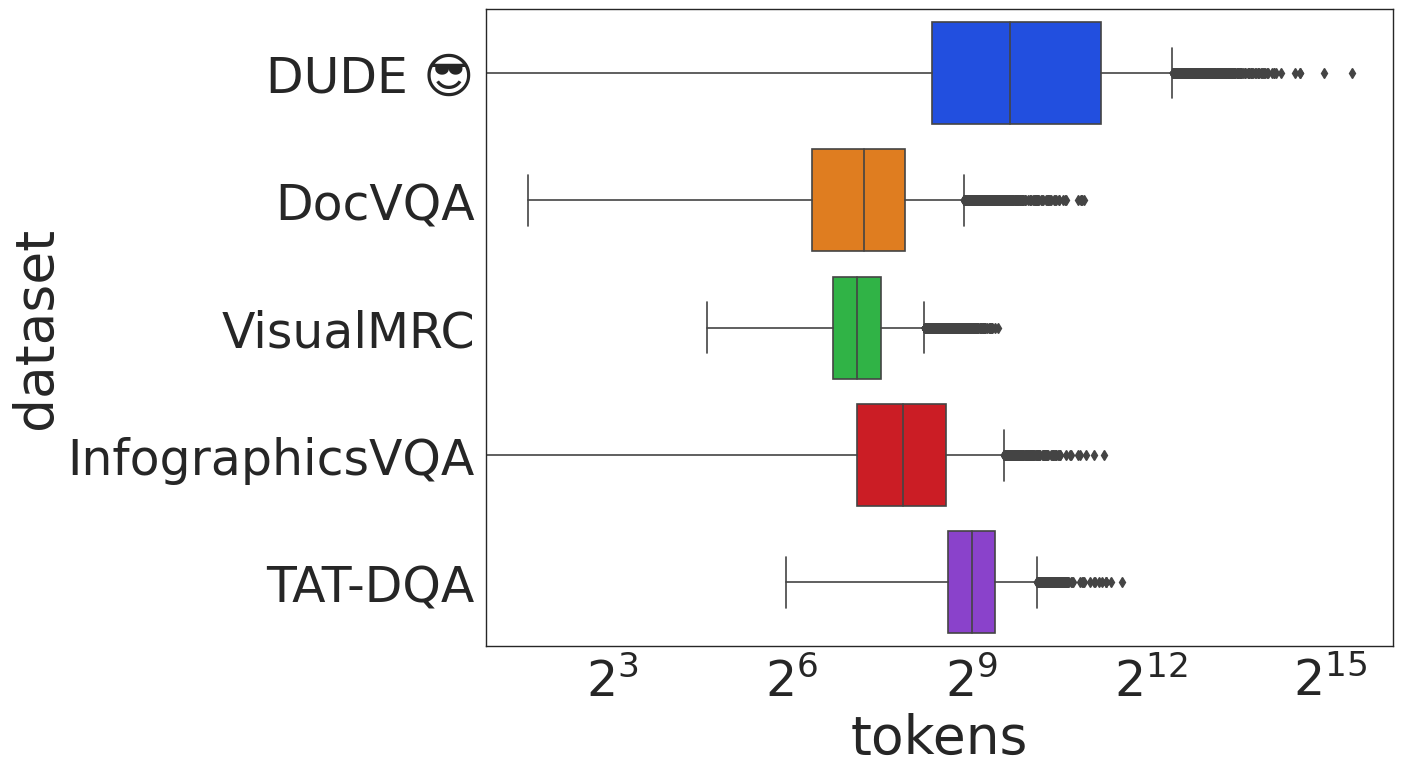}\hfill
\includegraphics[width=.28\linewidth,trim={12cm 0 0 0},clip]{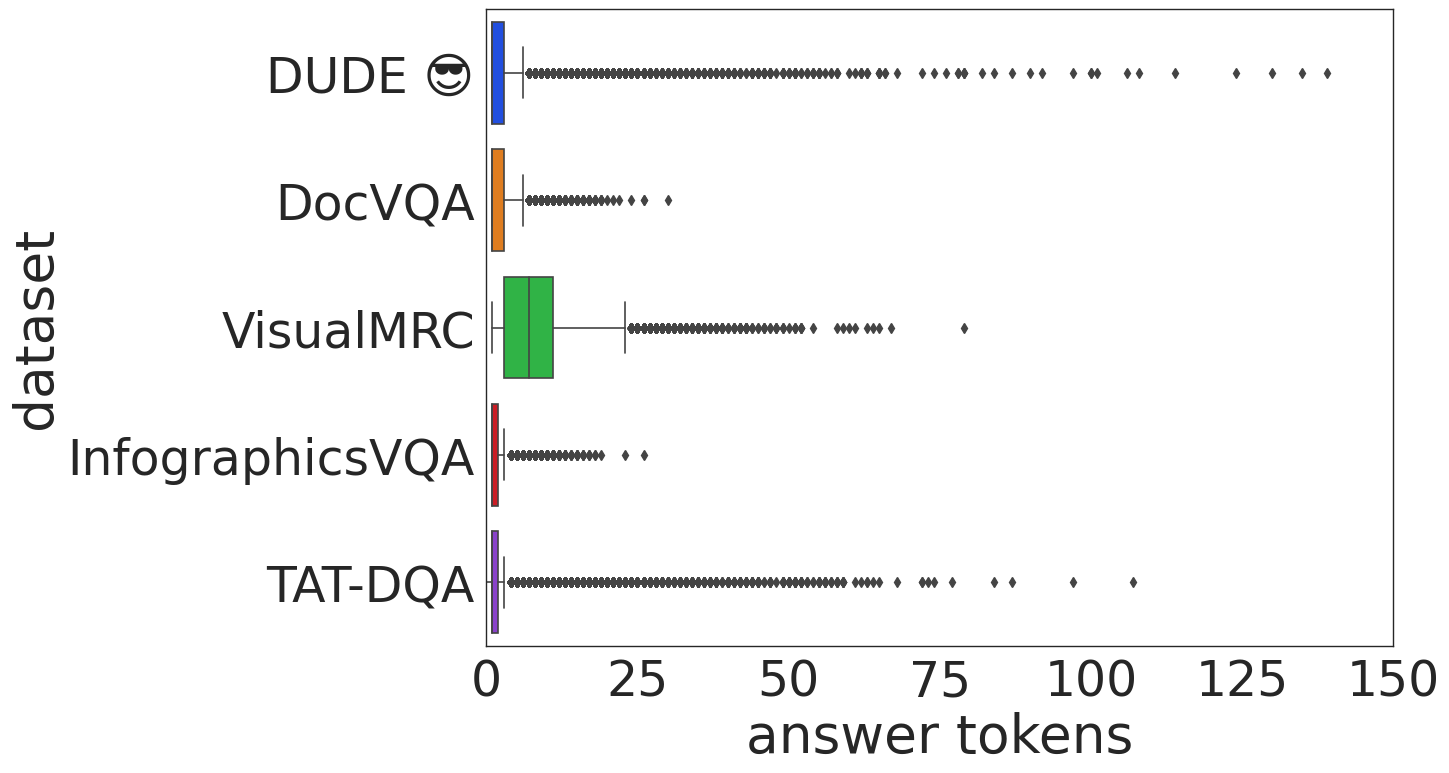}\hfill
\includegraphics[width=.275\linewidth,trim={12cm 0 0 0},clip]{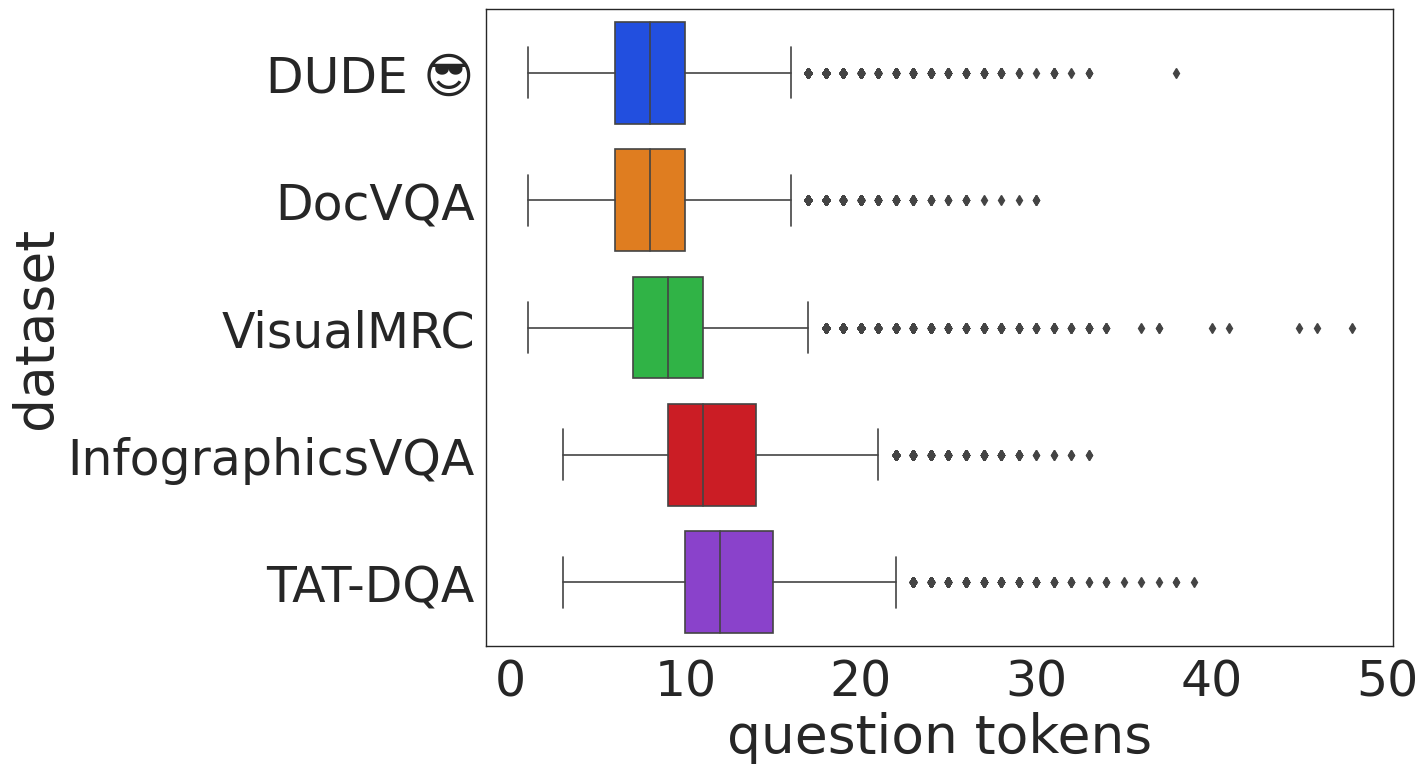}
\caption{Distribution of the number of tokens in documents, answers, and questions.}
\label{fig:token_ans_qn_dist}
\end{figure*}

\begin{figure}[h]
    \centering
\includegraphics[width=.8\linewidth]{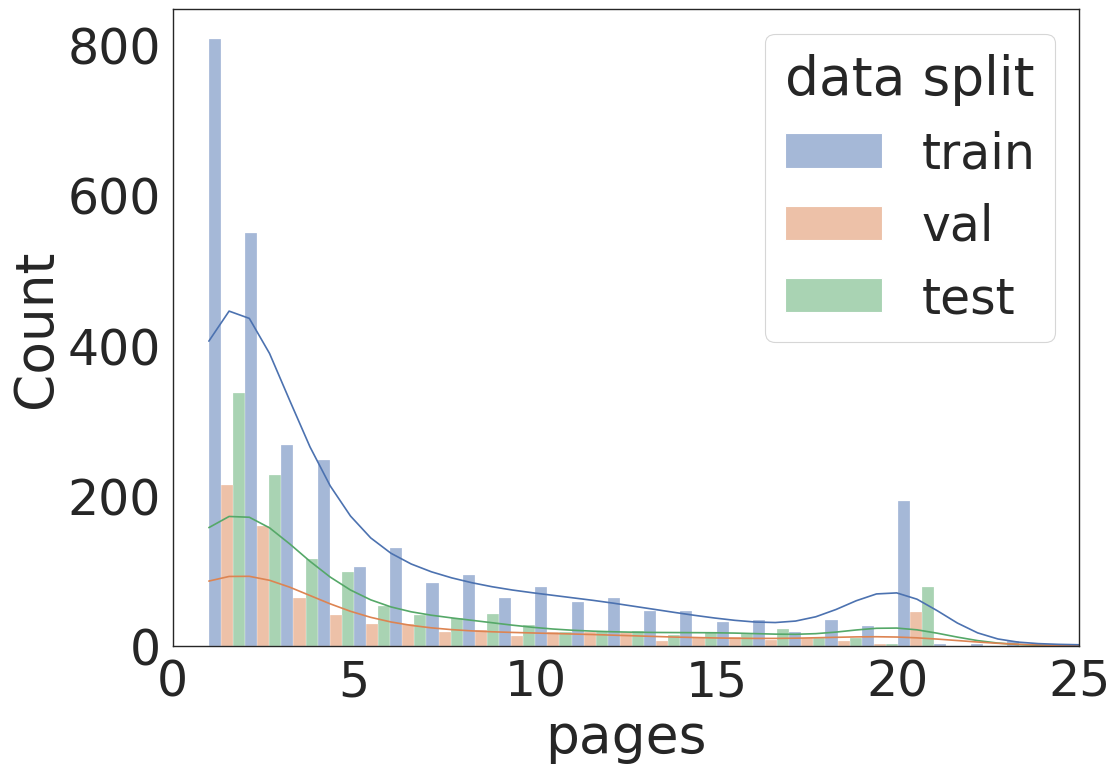}
    \caption{While other datasets are predominantly single-page only, the number of pages featuring in \project{} is more diverse, yet still biased towards shorter documents.
    }
    \label{fig:answer_lengths}
\end{figure}


\looseness=-1 The number of tokens in the document distribution is much more diverse compared to other datasets, a consequence of the more diverse distribution of pages (see Figure~\ref{fig:answer_lengths}). Note some of the documents are more visual than textual (or even visual-only), making the left whisker essentially reach 0 (\(\log_2\)-scaling of \(x\)-axis).

The distribution of the number of tokens in answers is heavy-tailed, to some extent this is also the property of the distribution of number of tokens in questions. Furthermore, 90.9\% of questions are unique, and so are 70.7\% of answers (taking answer variants into account).

We scrutinized the answer types by aggregating possible answers into classes representing the information they conveyed. 
The study used heuristics to determine if the answers fit into NER labeling scheme~\cite{SpaCyNER} or categories we anticipated, such as \emph{yes/no} and \emph{none}, or did not anticipate, such as \emph{color}.
This resulted in 25 different groups of answers, with the \emph{other} answer type being the fourth largest group.
Cramer's V coefficient was used to check for correlations between question types and answer types, and the results indicated that there were few correlations (see Appendix~\ref{app:correlation_heatmap}). The expected correlations, such as \textit{none} answers with \textit{not-answerable} questions or \textit{yes/no} answers with \textit{abstractive} questions, were present, but barely any correlation was significant. This suggests it is hard to guess the answer based on the question solely. 

We study relative diversity measure, called Simpson coefficient \cite{PAMI_places,SIMPSON_1949}. To define it, consider a fixed distance function \(d(a_1, a_2)\) defined for pair of documents \(a_1, a_2\in A\): the dataset. In our applications, it is the cosine similarity of a document embedding. Further, for an arbitrary number of datasets $A_1, \dots, A_N$ the diversity of $A_1$ with respect to $A_2, \dots, A_N$ is defined as 
\begin{equation*}
\mathrm{Div}_{A_2, \dots, A_N}(A_1) =
    1 - p\Bigl(d(a_{11}, a_{12}) < \min_{i=2:N} d(a_{i1}, a_{i2})  \Bigr)
\end{equation*}
where \(a_{i1},a_{i2}\in A_i\), are randomly selected, $i=2:Ni=2:N$. We report relative diversities of each of the datasets, relative to other datasets in the study, based on two embeddings: visual (ResNet-101 embeddings-based) and semantic (Tf-Idf embeddings-based), in Table~\ref{tab:dataset_summary}. The results show that the probability that two random documents from \project{} are more similar than each random pair of documents from other datasets is small
, meaning that documents in our dataset are well-distributed and diverse.
\subsection{Diagnostic Subsets}\label{sec:diagnostic}

\begin{figure}
    \centering
    \includegraphics[width=0.9\linewidth]{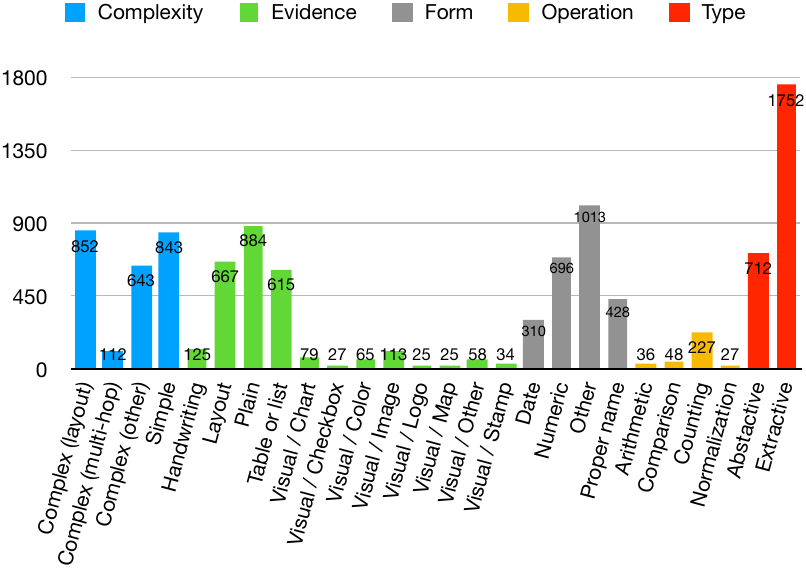}
    \caption{Count of particular diagnostic categories in a subset of 2.5k test set QA pairs annotated in detail to help analyze models' performance.}
    \label{fig:diagnostic}
\end{figure}

\looseness=-1 Following previous DU datasets, we gather diagnostic metadata for close to half of the documents and QA pairs in the test set (see \Cref{fig:diagnostic}). These are intended to enable a fine-grained analysis of the models' performance. The taxonomy used is an extension 
 of the one from earlier works \cite{mathew2020document,mathew2022infographicvqa,borchmann2021due}, covering \project-specific questions and enables a more detailed examination of visual artifacts under consideration.

\paragraph{Question type and perceived complexity.} We distinguish questions perceived as \textit{simple}, i.e., those based on spotting value near a phrase mentioned explicitly as a part of the question. For example, "Who is the Secretary of the U.S. Department of Commerce?" when the document contains "Penny Pritzker, Secretary, U.S. Department of Commerce." Such could be guessed given an approximate string matching algorithm and does not require much comprehension beyond that. The remaining questions are marked as \textit{hard} with distinguished categories of \textit{hard multi-hop questions}, and \textit{hard meta/layout-navigating questions}.

\paragraph{Answer evidence.} We provide information on what types of elements have to be comprehended to provide an answer, including \textit{free text}, \textit{handwriting}, \textit{table or list}, and \textit{layout}, i.e., non-tabular spatial understanding of text placement. These follow the ontology established by previous works \cite{mathew2020document,mathew2022infographicvqa,borchmann2021due}. In addition, we supply hints on graphical artifacts one needs to consider for particular questions, such as \textit{image/photo}, \textit{plot/chart}, \textit{checkbox}, and  \textit{annotation}.

\paragraph{Required operation.} We distinguish \textit{arithmetic}, \textit{comparison}, \textit{counting}, and \textit{normalization} operations to provide information on the need for performing, respectively, arithmetic operations on extractable data, comparing numerical values or sizes, counting elements or converting data present in the document to another format (e.g., rounding or date format conversion).

\paragraph{Answer form/shape.} Finally, we provide information on the shallow form of the returned answer, including \textit{date}, \textit{numeric}, and \textit{proper name}.

\subsection{Evaluation}\label{sec:evaluation}

The evaluation process follows the typical paradigm of separate training, validation, and test splits. We provide both a standalone evaluator and a website\footnote{\url{rrc.cvc.uab.es/?ch=23}} \cite{dude2023icdar} to submit test set predictions.

\looseness=-1 To assess models' performance, we rely on the ANLS metric introduced by authors of the ST-VQA dataset~\cite{biten2019scene}. Roughly speaking, it is a generalization of accuracy that does not penalize the system for an answer whose similarity to the gold standard measured with normalized Levenshtein similarity is above a specified threshold. Moreover, the metric assumes the presence of multiple, equally valid reference answers. The mentioned properties account for possible OCR errors or different phrasings, such as the same numerical answer represented as \textit{two} and \textit{2} by different annotators.

In practice, production DU systems provide an estimation of confidence in order to triage documents that do not need to be manually reviewed by a human. While the reliability of the automation ability of a DU solution is deemed quintessential for generating business value in practice \cite{bornet2021intelligent}, DU research rarely reports any confidence evaluation. Some exceptions are in closely related task domains like scene text recognition \cite{slossberg2020calibration} and QA \cite{kamath2020selective,zhang2021knowing}.



With DUDE, we want to establish calibration evaluation and confidence ranking as a default evaluation methodology in DU, especially since the field is so close to applications. 

\looseness=-1 To this end, we report (next to ANLS) two additional metrics, Expected Calibration Error ($\mathrm{ECE}$) \cite{niculescu2005predicting,naeini2015obtaining,guo2017calibration}, and Area\--Under\--Risk\--Coverage\--Curve ($\mathrm{AURC}$) \cite{geifman2017selective,jaeger2023a}.

Calibration requires that the probability a model assigns to its predictions equals their true likelihood of being correct \cite{dawid1982well,degroot1983comparison,zadrozny2002transforming}. 

$\mathrm{ECE}$ approximates top-1 calibration error by a weighted average over the accuracy/confidence difference of histogram bins. Particularly in our evaluation setting, we consider a predicted answer correct if its ANLS to the ground truth answer is above a pre-defined threshold ($\tau$=0.5). For consistency, not-answerable and list-answers both have confidence estimated for the answer as a whole (regardless of the number of answers). 
Following \cite{nixon2019measuring}, we apply equal-size binning (with 100 bins, $\mathcal{L}_p norm=1$), avoiding some pathologies of equal-range binning \cite{kumar2019calibration,vaicenavicius2019evaluating}. 

$\mathrm{AURC}$ is a selective classification metric that evaluates how well an estimator prevents silent failures on an \textit{i.i.d} test set. As an aggregate measure of estimator performance (ANLS) and confidence ranking, it provides a more practically useful estimate of overall performance when the estimator can abstain from (low-confidence) decisions and defer to a human for feedback.

By reporting the above metrics, we hope that in future work there will be contributions (e.g., calibration methods for improved forecasting or metrics for better predictive uncertainty evaluation) that concretely target the empirical observations of overconfidence/miscalibration in DU models.

\subsection{Baselines}

\paragraph{Human performance.} To establish the human baseline, we assign test set questions to \textit{Qualified Linguists}, ensuring none of them will face the same documents as reviewed in Phase 4. The procedure results in an estimation of 74.76 ANLS points (Table~\ref{res:baselines}). At first glance, this result seems low. Still, when analyzing results case by case, it turns out that it's hard to score much better since the answer format can influence the overall results a lot: \textit{Eagle} vs. \textit{an eagle} ($0.625$ ANLS), \textit{62\%} vs. \textit{62} (0.67 ANLS), \textit{1958-04-29} vs. \textit{4-29-58} (0 ANLS), \textit{Clemson University, Clemson South Carolina} vs. \textit{Clemson University} (0 ANLS). We achieved the lowest performance ($67.58$) on the extractive question type, which confirms our hypothesis since the abstractive answers are shorter (mostly numbers, yes/no, or colors).

\begin{figure*}[h]
    \centering
\includegraphics[width=0.59\linewidth]{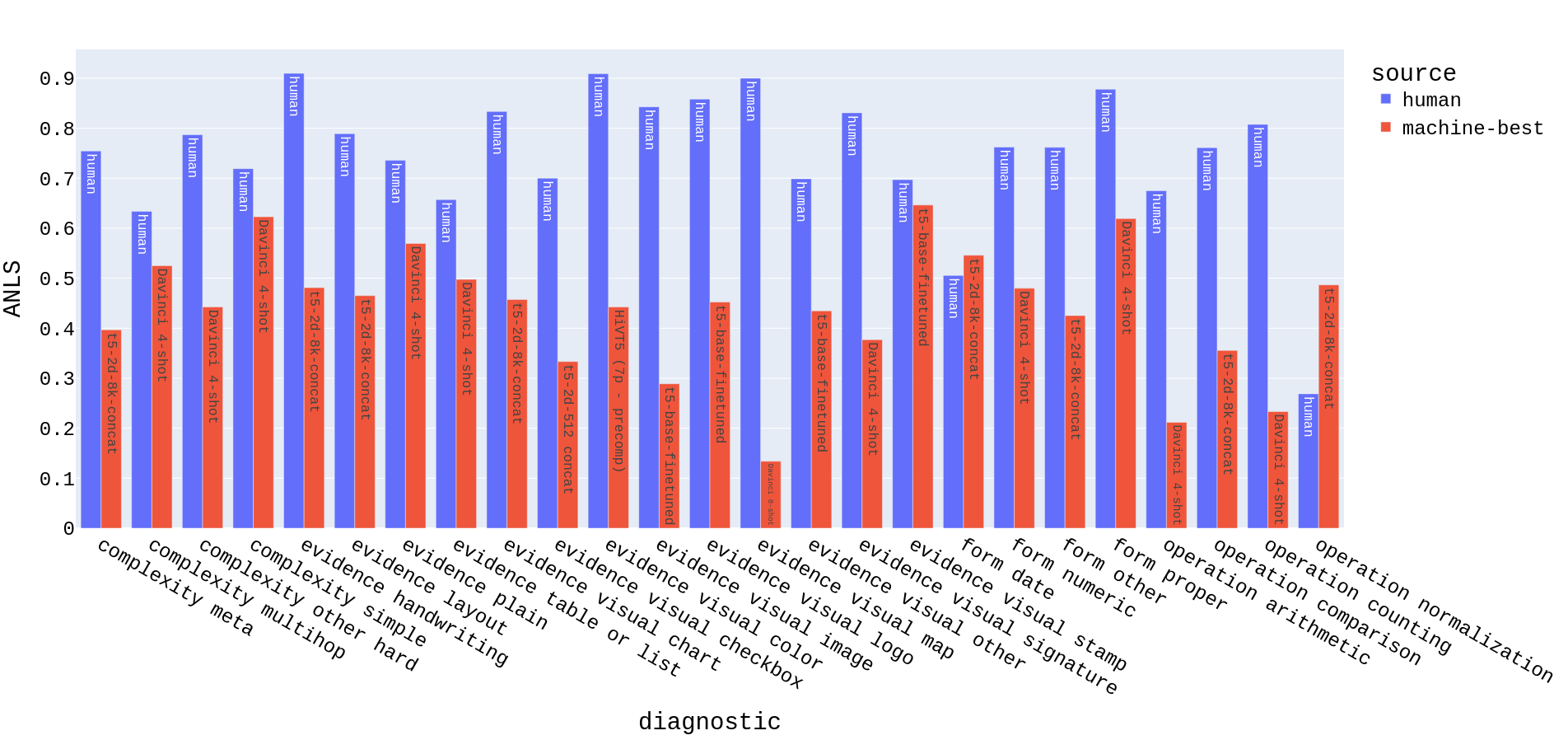}
    \caption{We report the average ANLS for the human expert vs. the best-performing model per diagnostic category as a ceiling analysis.} 
    \label{fig:ceiling-analysis}
\end{figure*}

We analyzed the maximum score achieved by the best-performing model for each diagnostic test category and plotted that against the human performance in Figure~\ref{fig:ceiling-analysis}.

\paragraph{Reference models.} We assessed a group of models
to determine how their performance is influenced by different factors such as (1) their ability to handle textual, layout, and visual elements, (2) whether they were fine-tuned for the task, (3) their size in (trainable parameters), and (4) the maximum input length they can handle.

To analyze factors (1) and (2), we conducted a zero-shot evaluation of several baseline text-only models. We used three encoder-based models (BERT \cite{devlin2018bert}, Longformer \cite{beltagy2020longformer}, and BigBird \cite{zaheer2020big}) that cannot generate text and three that feature a decoder (T5 \cite{raffel2020exploring}, GPT-3-Davinci \cite{brown2020language}, and ChatGPT) and have this capability. Next, we extended the T5 architecture with 2D layout embeddings \cite{borchmann2021due,Powalski2021GoingFB} and fine-tuned models with increasing maximum sequence lengths (512 $\to$ 8192) on \project{}. Finally, we evaluated our replication of the hierarchical Hi-VT5 model \cite{tito2022hierarchical}, as this model has the ability to decode text, understand multi-page layouts, and comprehend visual page features using DiT \cite{li2022dit}.

Regarding factors (2) and (3), we evaluated models of various sizes ranging from 131$\mathrm{M}$ (BigBird) to 175$\mathrm{B}$ (GPT-3-Davinci) and varied the input context from 512 (BERT) to 20480 (Hi-VT5) tokens.
Overall, we thoroughly evaluated multiple models in the different testing setups to determine their performance under various conditions, as seen in \Cref{res:baselines}.


\begin{table*}[t]
\centering
\begin{adjustbox}{width=1\textwidth}
\small
\bgroup
\def\arraystretch{0.95}
\begin{tabular}{lcccccccc||cccc}
\toprule Model &  Init. & Params & $\begin{array}{c}\text { Max Seq. } \\
\text { Length }\end{array}$ &  $\begin{array}{c}\text { Test } \\
\text { Setup }\end{array}$ & $\mathrm{ANLS_{all}}\uparrow$ & $\mathrm{ECE_{all}}\downarrow$ & $\mathrm{AURC_{all}}\downarrow$ & $\mathrm{ANLS_{do}}$ & $\begin{array}{c}\mathrm{ANLS_{do}} \\
\text { $\mathrm{Abs}$ }\end{array}$ & $\begin{array}{c}\text { $\mathrm{ANLS_{do}}$ } \\
\text { $\mathrm{Ex}$ }\end{array}$ & $\begin{array}{c}\text { $\mathrm{ANLS_{do}}$ } \\
\text { $\mathrm{NA}$ }\end{array}$ & $\begin{array}{c}\text { $\mathrm{ANLS_{do}}$ } \\
\text { $\mathrm{Li}$ }\end{array}$ \\ 

\midrule\multicolumn{13}{l}{\textit{text-only} Encoder-based models}\\ \midrule
Big Bird  & MPDocVQA & $131 \mathrm{M}$ & 4096 & Concat* & 26.27 & 30.14    & 44.22 & 30.67 & 7.11 & 40.26 & 12.75 & 8.46 \\ 
BERT-Large & MPDocVQA & $334 \mathrm{M}$ & 512 & Max Conf.* & 25.48 & 34.06 & 48.60 & 32.18 & 7.28 & 42.23 & 5.88 & 11.13 \\
Longformer & MPDocVQA & $148 \mathrm{M}$ & 4096 & Concat* & 27.14 & 27.59   & 44.59 & 33.45  & 8.55 & 43.58 & 10.78 & 10.62 \\

\midrule\multicolumn{13}{l}{\textit{text-only} Encoder-Decoder based models}\\ \midrule
T5 & base & $223 \mathrm{M}$ & 512 & Concat-0* & 19.65 & 19.14  & 48.83 & 25.62 & 5.24 & 33.91 & 0 & 7.31  \\
T5 & MPDocVQA & 223M & 512 & Max Conf.* & 29.48 & 27.18         & 43.06 & 37.56 & 21.19 & 44.22 & 0 & 10.56  \\ 
T5 & base & $223 \mathrm{M}$ & 512 & Concat+FT & 37.41 & \textbf{10.82} & 41.09 & 40.61 & 42.61 & 48.20 & 53.92 & 16.87\\
T5 & base & $223 \mathrm{M}$ & 8192 & Concat+FT & 41.80 & 17.33 & 49.53 & 44.95 & 
 47.62 & 50.49 & 63.72 & 7.56  \\

\midrule \multicolumn{13}{l}{\textit{text-only} Large Language models (LLM)}\\ \midrule
ChatGPT  & gpt-3.5-turbo & $20 \mathrm{B}$ & 4096 & Concat-0 & - & - & - & 35.07 & 16.73 & 42.52 & 70.59 & 15.97  \\
& & & & Concat-4 & - & - & - & 41.89 & 22.19 & 49.90 & \textbf{77.45} & 17.74  \\
GPT3 & davinci3 & $175 \mathrm{B}$ & 4000 & Concat-0 & - & - & - & 43.95 & 18.16 & 54.44 & 73.53 & 36.32  \\
& & & & Concat-4 & - & - & - & 47.04 &  22.37 & \textbf{57.09} & 63.73 & \textbf{40.01}  \\

\midrule\multicolumn{13}{l}{\textit{text+layout} Encoder-Decoder based models}\\ \midrule
T5-2D & base & $223 \mathrm{M}$ & 512 & Concat+FT & 37.10 & 10.85 & 41.46 & 40.50 & 42.48 & 48.62 & 52.94 & 3.49  \\
T5-2D & base & $223 \mathrm{M}$ & 8192 & Concat+FT & 42.10 & 17.00  & 48.83 & 45.73 & 48.37 & 52.29 & 63.72 & 8.02\\
T5-2D & large & $770 \mathrm{M}$ & 8192 & Concat+FT & \textbf{46.06} & 14.40 & \textbf{35.70} & \textbf{48.14} & \textbf{50.81} & 55.65 & 68.62 & 5.43  \\

\midrule \multicolumn{13}{l}{\textit{text+layout+vision} models} \\ \midrule
HiVT5  &  & 316M & 20480 & Hierarchical+FT & 23.06 & 11.91                   & 54.35 & 22.33 & 33.94 & 17.60 & 61.76 & 6.83  \\
LayoutLMv3  & MPDocVQA & $125 \mathrm{M}$ & 512 & Max Conf.* & 20.31 & 34.97 & 47.51 & 25.27 & 8.10 & 32.60 & 8.82 & 7.82 \\
\midrule
\midrule
\textit{Human} baseline  &  & & & & & & & 74.76 & 81.95 & 67.58 & 83.33 & 67.74 \\
\bottomrule
\end{tabular}
\egroup
\end{adjustbox}
\caption{\jordyskeleton{\looseness=-1 Summary of Baseline performance on the \project{} test set ($_{all}$) and diagnostic subset ($_{do}$). 
Test setups are defined as \textit{Max Conf.}: predict one answer per page and return an answer with the highest probability over all pages, \textit{Concat}: predict on tokens truncated to maximum sequence length, \textit{FT} stands for fine-tuning on \project{} training data, and \textit{-0} refers to zero-shot and \textit{-4} few-shot inference.
Average ANLS results per question type are abbreviated as (Abs)tractive, (Ex)tractive, (N)ot-(A)nswerable, (Li)st. (*) We report only results for best performing test setup (either \textit{Max Conf.} or \textit{Concat}). All scalars are scaled between 0 and 100 for readability.} 
\label{res:baselines}}
\end{table*}

\subsection{Analysis \& Discussion}


\pietruszkacolor{ 
To summarize, our study reveals that existing advanced language models such as BERT, Longformer, and BigBird struggle with comprehending visual elements and document layouts. To address this issue, we introduced T5, T5-2D, and Hi-VT5 models that incorporate layout and visual information. Still, their performance remains unsatisfactory, as evidenced by the comparison with the human baseline, similar to what has been reported for InfographicsVQA. This indicates that there is still scope for enhancing the visual understanding of \project{} models. Moreover, our findings indicate that a large LLM capable of processing long inputs alone is insufficient for achieving strong performance in \project{}, especially for the extractive type of answer. Finally, the dataset's length significantly affects the models' scores, as seen by the increase in scores by $4.4-5.0$ points when the T5 and T5+2D context length is extended from 512 to 8192.
Similarly, the model size has a positive correlation with the final score, but it holds only within a particular model-type and is not the main factor influencing the results.
State-of-the-art performance of $46.04$ $\mathrm{ANLS_{all}}$ was achieved on $T5_{large}$ with a 2D layout understanding that consumed 8192 tokens, confirming the observation above.
}

\section{Conclusion}
\pietruszkacolor{
In conclusion, this paper introduces a new large-scale multi-paged, multi-domain, multi-industry Document Visual Question Answering Benchmark for document understanding. Our dataset is adjusted to the real-world environment where we need to process long documents and understand different types of documents. The benchmark includes visual semantics such as \textit{tables, charts, figures, lists, checkboxes, stamps}, and more, which are essential for real-world document understanding.
The performance of state-of-the-art textual and multi-modal models still lags behind human performance, indicating the need for further improvement in visual understanding for DU models. Nevertheless, we believe evaluating systems on \project{} could inspire new architectures and methods.
}
\paragraph{Limitations.} \label{sec:future}
\pietruszkacolor{
As our approach is closer to real-world industrial applications, and enables models to recognize and understand new unseen data without the need for re-training, it does come with some limitations and constraining factors, including the use of only English language documents. Future work could address these limitations and expand the benchmark to include other languages. Moreover, although our dataset can be considered large-scale, it still represents a relatively small sample size of the plethora of documents that exist in the real world.
}

{\small
\bibliographystyle{ieee_fullname}
\bibliography{main}

\begin{thebibliography}{100}\itemsep=-1pt

\bibitem{SpaCyNER}
{SpaCy} \texttt{en\_core\_web\_lg} label scheme.
\newblock \url{https://spacy.io/models/en}.
\newblock Accessed: 2023-03-08.

\bibitem{agashe-etal-2019-juice}
Rajas Agashe, Srinivasan Iyer, and Luke Zettlemoyer.
\newblock {J}u{IC}e: A large scale distantly supervised dataset for open domain
  context-based code generation.
\newblock In {\em Proceedings of the 2019 Conference on Empirical Methods in
  Natural Language Processing and the 9th International Joint Conference on
  Natural Language Processing (EMNLP-IJCNLP)}, pages 5436--5446, Hong Kong,
  China, Nov. 2019. Association for Computational Linguistics.

\bibitem{https://doi.org/10.48550/arxiv.1505.00468}
Aishwarya Agrawal, Jiasen Lu, Stanislaw Antol, Margaret Mitchell, C.~Lawrence
  Zitnick, Dhruv Batra, and Devi Parikh.
\newblock Vqa: Visual question answering, 2015.

\bibitem{https://doi.org/10.48550/arxiv.1905.13319}
Aida Amini, Saadia Gabriel, Peter Lin, Rik Koncel-Kedziorski, Yejin Choi, and
  Hannaneh Hajishirzi.
\newblock Mathqa: Towards interpretable math word problem solving with
  operation-based formalisms, 2019.

\bibitem{appalaraju2021docformer}
Srikar Appalaraju, Bhavan Jasani, Bhargava~Urala Kota, Yusheng Xie, and R
  Manmatha.
\newblock Docformer: End-to-end transformer for document understanding.
\newblock In {\em Proceedings of the IEEE/CVF International Conference on
  Computer Vision}, pages 993--1003, 2021.

\bibitem{beltagy2020longformer}
Iz Beltagy, Matthew~E Peters, and Arman Cohan.
\newblock Longformer: The long-document transformer.
\newblock {\em arXiv preprint arXiv:2004.05150}, 2020.

\bibitem{biten2019icdar}
Ali~Furkan Biten, Ruben Tito, Andres Mafla, Lluis Gomez, Mar{\c{c}}al Rusinol,
  Minesh Mathew, CV Jawahar, Ernest Valveny, and Dimosthenis Karatzas.
\newblock Icdar 2019 competition on scene text visual question answering.
\newblock In {\em 2019 International Conference on Document Analysis and
  Recognition (ICDAR)}, pages 1563--1570. IEEE, 2019.

\bibitem{biten2019scene}
Ali~Furkan Biten, Ruben Tito, Andres Mafla, Lluis Gomez, Mar{\c{c}}al Rusinol,
  Ernest Valveny, CV Jawahar, and Dimosthenis Karatzas.
\newblock Scene text visual question answering.
\newblock In {\em Proceedings of the IEEE/CVF international conference on
  computer vision}, 2019.

\bibitem{bjerva-etal-2020-subjqa}
Johannes Bjerva, Nikita Bhutani, Behzad Golshan, Wang-Chiew Tan, and Isabelle
  Augenstein.
\newblock {SubjQA}: {A} {D}ataset for {S}ubjectivity and {R}eview
  {C}omprehension.
\newblock In {\em Proceedings of the 2020 Conference on Empirical Methods in
  Natural Language Processing (EMNLP)}, pages 5480--5494, Online, Nov. 2020.
  Association for Computational Linguistics.

\bibitem{borchmann2021due}
{\L}ukasz Borchmann, Micha{\l} Pietruszka, Tomasz Stanislawek, Dawid
  Jurkiewicz, Micha{\l} Turski, Karolina Szyndler, and Filip Grali{\'n}ski.
\newblock Due: End-to-end document understanding benchmark.
\newblock In {\em Thirty-fifth Conference on Neural Information Processing
  Systems Datasets and Benchmarks Track (Round 2)}, 2021.

\bibitem{bornet2021intelligent}
Pascal Bornet, Ian Barkin, and Jochen Wirtz.
\newblock {\em Intelligent automation: Welcome to the world of hyperautomation:
  learn how to harness artificial intelligence to boost business \& make our
  world more human}.
\newblock World Scientific, 2021.

\bibitem{brown2020language}
Tom Brown, Benjamin Mann, Nick Ryder, Melanie Subbiah, Jared~D Kaplan, Prafulla
  Dhariwal, Arvind Neelakantan, Pranav Shyam, Girish Sastry, Amanda Askell,
  et~al.
\newblock Language models are few-shot learners.
\newblock {\em Advances in neural information processing systems},
  33:1877--1901, 2020.

\bibitem{castro-etal-2020-lifeqa}
Santiago Castro, Mahmoud Azab, Jonathan Stroud, Cristina Noujaim, Ruoyao Wang,
  Jia Deng, and Rada Mihalcea.
\newblock {L}ife{QA}: A real-life dataset for video question answering.
\newblock In {\em Proceedings of the Twelfth Language Resources and Evaluation
  Conference}, pages 4352--4358, Marseille, France, May 2020. European Language
  Resources Association.

\bibitem{castro-etal-2022-wild}
Santiago Castro, Naihao Deng, Pingxuan Huang, Mihai Burzo, and Rada Mihalcea.
\newblock In-the-wild video question answering.
\newblock In {\em Proceedings of the 29th International Conference on
  Computational Linguistics}, pages 5613--5635, Gyeongju, Republic of Korea,
  Oct. 2022. International Committee on Computational Linguistics.

\bibitem{https://doi.org/10.48550/arxiv.2211.08545}
Shuaichen Chang, David Palzer, Jialin Li, Eric Fosler-Lussier, and Ningchuan
  Xiao.
\newblock Mapqa: A dataset for question answering on choropleth maps, 2022.

\bibitem{chen-etal-2021-geoqa}
Jiaqi Chen, Jianheng Tang, Jinghui Qin, Xiaodan Liang, Lingbo Liu, Eric Xing,
  and Liang Lin.
\newblock {G}eo{QA}: A geometric question answering benchmark towards
  multimodal numerical reasoning.
\newblock In {\em Findings of the Association for Computational Linguistics:
  ACL-IJCNLP 2021}, pages 513--523, Online, Aug. 2021. Association for
  Computational Linguistics.

\bibitem{colas-etal-2020-tutorialvqa}
Anthony Colas, Seokhwan Kim, Franck Dernoncourt, Siddhesh Gupte, Zhe Wang, and
  Doo~Soon Kim.
\newblock {T}utorial{VQA}: Question answering dataset for tutorial videos.
\newblock In {\em Proceedings of the Twelfth Language Resources and Evaluation
  Conference}, pages 5450--5455, Marseille, France, May 2020. European Language
  Resources Association.

\bibitem{dasigi2019quoref}
Pradeep Dasigi, Nelson~F Liu, Ana Marasovi{\'c}, Noah~A Smith, and Matt
  Gardner.
\newblock Quoref: A reading comprehension dataset with questions requiring
  coreferential reasoning.
\newblock {\em arXiv preprint arXiv:1908.05803}, 2019.

\bibitem{dawid1982well}
A~Philip Dawid.
\newblock The well-calibrated bayesian.
\newblock {\em Journal of the American Statistical Association},
  77(379):605--610, 1982.

\bibitem{degroot1983comparison}
Morris~H DeGroot and Stephen~E Fienberg.
\newblock The comparison and evaluation of forecasters.
\newblock {\em Journal of the Royal Statistical Society: Series D (The
  Statistician)}, 32(1-2):12--22, 1983.

\bibitem{devlin2018bert}
Jacob Devlin, Ming-Wei Chang, Kenton Lee, and Kristina Toutanova.
\newblock Bert: Pre-training of deep bidirectional transformers for language
  understanding.
\newblock In {\em Proceedings of the 2019 Conference of the North American
  Chapter of the Association for Computational Linguistics: Human Language
  Technologies, Volume 1 (Long and Short Papers)}, pages 4171--4186, 2019.

\bibitem{dutt-etal-2022-perkgqa}
Ritam Dutt, Kasturi Bhattacharjee, Rashmi Gangadharaiah, Dan Roth, and Carolyn
  Rose.
\newblock {P}er{KGQA}: Question answering over personalized knowledge graphs.
\newblock In {\em Findings of the Association for Computational Linguistics:
  NAACL 2022}, pages 253--268, Seattle, United States, July 2022. Association
  for Computational Linguistics.

\bibitem{Garncarek2021LAMBERTLL}
Lukasz Garncarek, Rafal Powalski, Tomasz Stanislawek, Bartosz Topolski, Piotr
  Halama, Michał Turski, and Filip Graliński.
\newblock Lambert: Layout-aware language modeling using bert for information
  extraction.
\newblock In {\em ICDAR}, 2021.

\bibitem{geifman2017selective}
Yonatan Geifman and Ran El-Yaniv.
\newblock Selective classification for deep neural networks.
\newblock {\em Advances in neural information processing systems}, 30, 2017.

\bibitem{gu2021unidoc}
Jiuxiang Gu, Jason Kuen, Vlad~I Morariu, Handong Zhao, Rajiv Jain, Nikolaos
  Barmpalios, Ani Nenkova, and Tong Sun.
\newblock Unidoc: Unified pretraining framework for document understanding.
\newblock {\em Advances in Neural Information Processing Systems}, 34:39--50,
  2021.

\bibitem{gui-etal-2017-question}
Lin Gui, Jiannan Hu, Yulan He, Ruifeng Xu, Qin Lu, and Jiachen Du.
\newblock A question answering approach for emotion cause extraction.
\newblock In {\em Proceedings of the 2017 Conference on Empirical Methods in
  Natural Language Processing}, pages 1593--1602, Copenhagen, Denmark, Sept.
  2017. Association for Computational Linguistics.

\bibitem{guo2017calibration}
Chuan Guo, Geoff Pleiss, Yu Sun, and Kilian~Q. Weinberger.
\newblock On calibration of modern neural networks.
\newblock In {\em Proceedings of the 34th International Conference on Machine
  Learning - Volume 70}, ICML'17, page 1321–1330, 2017.

\bibitem{gupta-demner-fushman-2022-overview}
Deepak Gupta and Dina Demner-Fushman.
\newblock Overview of the {M}ed{V}id{QA} 2022 shared task on medical video
  question-answering.
\newblock In {\em Proceedings of the 21st Workshop on Biomedical Language
  Processing}, pages 264--274, Dublin, Ireland, May 2022. Association for
  Computational Linguistics.

\bibitem{https://doi.org/10.48550/arxiv.1802.08218}
Danna Gurari, Qing Li, Abigale~J. Stangl, Anhong Guo, Chi Lin, Kristen Grauman,
  Jiebo Luo, and Jeffrey~P. Bigham.
\newblock Vizwiz grand challenge: Answering visual questions from blind people,
  2018.

\bibitem{harley2015evaluation}
Adam~W Harley, Alex Ufkes, and Konstantinos~G Derpanis.
\newblock Evaluation of deep convolutional nets for document image
  classification and retrieval.
\newblock In {\em 2015 13th International Conference on Document Analysis and
  Recognition (ICDAR)}, pages 991--995. IEEE, 2015.

\bibitem{hopkins-etal-2019-semeval}
Mark Hopkins, Ronan Le~Bras, Cristian Petrescu-Prahova, Gabriel Stanovsky,
  Hannaneh Hajishirzi, and Rik Koncel-Kedziorski.
\newblock {S}em{E}val-2019 task 10: Math question answering.
\newblock In {\em Proceedings of the 13th International Workshop on Semantic
  Evaluation}, pages 893--899, Minneapolis, Minnesota, USA, June 2019.
  Association for Computational Linguistics.

\bibitem{hu-etal-2022-chef}
Xuming Hu, Zhijiang Guo, GuanYu Wu, Aiwei Liu, Lijie Wen, and Philip Yu.
\newblock {CHEF}: A pilot {C}hinese dataset for evidence-based fact-checking.
\newblock In {\em Proceedings of the 2022 Conference of the North American
  Chapter of the Association for Computational Linguistics: Human Language
  Technologies}, pages 3362--3376, Seattle, United States, July 2022.
  Association for Computational Linguistics.

\bibitem{huang2022layoutlmv3}
Yupan Huang, Tengchao Lv, Lei Cui, Yutong Lu, and Furu Wei.
\newblock Layoutlmv3: Pre-training for document ai with unified text and image
  masking.
\newblock {\em arXiv preprint arXiv:2204.08387}, 2022.

\bibitem{jaeger2023a}
Paul~F Jaeger, Carsten~Tim L{\"u}th, Lukas Klein, and Till~J. Bungert.
\newblock A call to reflect on evaluation practices for failure detection in
  image classification.
\newblock In {\em International Conference on Learning Representations}, 2023.

\bibitem{jaume2019funsd}
Guillaume Jaume, Hazim~Kemal Ekenel, and Jean-Philippe Thiran.
\newblock Funsd: A dataset for form understanding in noisy scanned documents.
\newblock In {\em 2019 International Conference on Document Analysis and
  Recognition Workshops (ICDARW)}, volume~2, pages 1--6. IEEE, 2019.

\bibitem{jin-etal-2019-pubmedqa}
Qiao Jin, Bhuwan Dhingra, Zhengping Liu, William Cohen, and Xinghua Lu.
\newblock {P}ub{M}ed{QA}: A dataset for biomedical research question answering.
\newblock In {\em Proceedings of the 2019 Conference on Empirical Methods in
  Natural Language Processing and the 9th International Joint Conference on
  Natural Language Processing (EMNLP-IJCNLP)}, pages 2567--2577, Hong Kong,
  China, Nov. 2019. Association for Computational Linguistics.

\bibitem{kacupaj-etal-2021-conversational}
Endri Kacupaj, Joan Plepi, Kuldeep Singh, Harsh Thakkar, Jens Lehmann, and
  Maria Maleshkova.
\newblock Conversational question answering over knowledge graphs with
  transformer and graph attention networks.
\newblock In {\em Proceedings of the 16th Conference of the European Chapter of
  the Association for Computational Linguistics: Main Volume}, pages 850--862,
  Online, Apr. 2021. Association for Computational Linguistics.

\bibitem{kamath2020selective}
Amita Kamath, Robin Jia, and Percy Liang.
\newblock Selective question answering under domain shift.
\newblock In {\em Proceedings of the 58th Annual Meeting of the Association for
  Computational Linguistics}, pages 5684--5696, 2020.

\bibitem{kamath:hal-01759306}
Sanjay Kamath, Brigitte Grau, and Yue Ma.
\newblock {Verification of the Expected Answer Type for Biomedical Question
  Answering}.
\newblock In {\em {First International Workshop on Hybrid Question Answering
  with Structured and Unstructured Knowledge (HQA'18)}}, WWW '18 Companion
  Proceedings of the The Web Conference 2018, pages 1093--1097, Lyon, France,
  Apr. 2018. {ACM Press}.

\bibitem{projectid}
Andreas Kirsch.
\newblock Player of jeopardy: Chatgpt evaluation, 2023.

\bibitem{kumar2019calibration}
Ananya Kumar, Percy Liang, and Tengyu Ma.
\newblock Verified uncertainty calibration.
\newblock In {\em Advances in Neural Information Processing Systems}, 2019.

\bibitem{kwiatkowski2019natural}
Tom Kwiatkowski, Jennimaria Palomaki, Olivia Redfield, Michael Collins, Ankur
  Parikh, Chris Alberti, Danielle Epstein, Illia Polosukhin, Jacob Devlin,
  Kenton Lee, et~al.
\newblock Natural questions: a benchmark for question answering research.
\newblock {\em Transactions of the Association for Computational Linguistics},
  2019.

\bibitem{lakomkin-etal-2018-kt}
Egor Lakomkin, Sven Magg, Cornelius Weber, and Stefan Wermter.
\newblock {KT}-speech-crawler: Automatic dataset construction for speech
  recognition from {Y}ou{T}ube videos.
\newblock In {\em Proceedings of the 2018 Conference on Empirical Methods in
  Natural Language Processing: System Demonstrations}, pages 90--95, Brussels,
  Belgium, Nov. 2018. Association for Computational Linguistics.

\bibitem{lei-etal-2018-tvqa}
Jie Lei, Licheng Yu, Mohit Bansal, and Tamara Berg.
\newblock {TVQA}: Localized, compositional video question answering.
\newblock In {\em Proceedings of the 2018 Conference on Empirical Methods in
  Natural Language Processing}, pages 1369--1379, Brussels, Belgium, Oct.-Nov.
  2018. Association for Computational Linguistics.

\bibitem{levenshtein1966binary}
Vladimir~I Levenshtein et~al.
\newblock Binary codes capable of correcting deletions, insertions, and
  reversals.
\newblock In {\em Soviet physics doklady}, volume~10, pages 707--710. Soviet
  Union, 1966.

\bibitem{li2022multispanqa}
Haonan Li, Martin Tomko, Maria Vasardani, and Timothy Baldwin.
\newblock Multispanqa: A dataset for multi-span question answering.
\newblock In {\em Proceedings of the 2022 Conference of the North American
  Chapter of the Association for Computational Linguistics: Human Language
  Technologies}, pages 1250--1260, 2022.

\bibitem{li2022dit}
Junlong Li, Yiheng Xu, Tengchao Lv, Lei Cui, Cha Zhang, and Furu Wei.
\newblock Dit: Self-supervised pre-training for document image transformer.
\newblock In {\em Proceedings of the 30th ACM International Conference on
  Multimedia}, pages 3530--3539, 2022.

\bibitem{li-etal-2021-mlec}
Jing Li, Shangping Zhong, and Kaizhi Chen.
\newblock {MLEC-QA}: {A} {C}hinese {M}ulti-{C}hoice {B}iomedical {Q}uestion
  {A}nswering {D}ataset.
\newblock In {\em Proceedings of the 2021 Conference on Empirical Methods in
  Natural Language Processing}, pages 8862--8874, Online and Punta Cana,
  Dominican Republic, Nov. 2021. Association for Computational Linguistics.

\bibitem{li2020docbank}
Minghao Li, Yiheng Xu, Lei Cui, Shaohan Huang, Furu Wei, Zhoujun Li, and Ming
  Zhou.
\newblock Docbank: A benchmark dataset for document layout analysis, 2020.

\bibitem{li2021selfdoc}
Peizhao Li, Jiuxiang Gu, Jason Kuen, Vlad~I Morariu, Handong Zhao, Rajiv Jain,
  Varun Manjunatha, and Hongfu Liu.
\newblock Selfdoc: Self-supervised document representation learning.
\newblock In {\em Proceedings of the IEEE/CVF Conference on Computer Vision and
  Pattern Recognition}, pages 5652--5660, 2021.

\bibitem{liu-wan-2021-codeqa-question}
Chenxiao Liu and Xiaojun Wan.
\newblock {C}ode{QA}: A question answering dataset for source code
  comprehension.
\newblock In {\em Findings of the Association for Computational Linguistics:
  EMNLP 2021}, pages 2618--2632, Punta Cana, Dominican Republic, Nov. 2021.
  Association for Computational Linguistics.

\bibitem{ijcai2020p0501}
Jian Liu, Leyang Cui, Hanmeng Liu, Dandan Huang, Yile Wang, and Yue Zhang.
\newblock Logiqa: A challenge dataset for machine reading comprehension with
  logical reasoning.
\newblock In Christian Bessiere, editor, {\em Proceedings of the Twenty-Ninth
  International Joint Conference on Artificial Intelligence, {IJCAI-20}}, pages
  3622--3628. International Joint Conferences on Artificial Intelligence
  Organization, 7 2020.
\newblock Main track.

\bibitem{liu-etal-2019-xqa}
Jiahua Liu, Yankai Lin, Zhiyuan Liu, and Maosong Sun.
\newblock {XQA}: A cross-lingual open-domain question answering dataset.
\newblock In {\em Proceedings of the 57th Annual Meeting of the Association for
  Computational Linguistics}, pages 2358--2368, Florence, Italy, July 2019.
  Association for Computational Linguistics.

\bibitem{https://doi.org/10.48550/arxiv.2007.15207}
Shayne Longpre, Yi Lu, and Joachim Daiber.
\newblock Mkqa: A linguistically diverse benchmark for multilingual open domain
  question answering, 2020.

\bibitem{10.1007/978-3-031-06555-2_44}
Ibrahim~Souleiman Mahamoud, Micka{\"e}l Coustaty, Aur{\'e}lie Joseph,
  Vincent~Poulain d'Andecy, and Jean-Marc Ogier.
\newblock Qalayout: Question answering layout based on multimodal attention for
  visual question answering on corporate document.
\newblock In Seiichi Uchida, Elisa Barney, and V{\'e}ronique Eglin, editors,
  {\em Document Analysis Systems}, pages 659--673, Cham, 2022. Springer
  International Publishing.

\bibitem{mathew2022infographicvqa}
Minesh Mathew, Viraj Bagal, Rub{\`e}n Tito, Dimosthenis Karatzas, Ernest
  Valveny, and CV Jawahar.
\newblock Infographicvqa.
\newblock In {\em Proceedings of the IEEE/CVF Winter Conference on Applications
  of Computer Vision}, pages 1697--1706, 2022.

\bibitem{mathew2020document}
Minesh Mathew, Ruben Tito, Dimosthenis Karatzas, R Manmatha, and CV Jawahar.
\newblock Document visual question answering challenge 2020.
\newblock {\em arXiv preprint arXiv:2008.08899}, 2020.

\bibitem{https://doi.org/10.48550/arxiv.2004.10645}
Sewon Min, Julian Michael, Hannaneh Hajishirzi, and Luke Zettlemoyer.
\newblock Ambigqa: Answering ambiguous open-domain questions, 2020.

\bibitem{mishra2019ocr}
Anand Mishra, Shashank Shekhar, Ajeet~Kumar Singh, and Anirban Chakraborty.
\newblock Ocr-vqa: Visual question answering by reading text in images.
\newblock In {\em 2019 international conference on document analysis and
  recognition (ICDAR)}, pages 947--952. IEEE, 2019.

\bibitem{mishra-etal-2022-numglue}
Swaroop Mishra, Arindam Mitra, Neeraj Varshney, Bhavdeep Sachdeva, Peter Clark,
  Chitta Baral, and Ashwin Kalyan.
\newblock {N}um{GLUE}: A suite of fundamental yet challenging mathematical
  reasoning tasks.
\newblock In {\em Proceedings of the 60th Annual Meeting of the Association for
  Computational Linguistics (Volume 1: Long Papers)}, pages 3505--3523, Dublin,
  Ireland, May 2022. Association for Computational Linguistics.

\bibitem{moller-etal-2020-covid}
Timo M{\"o}ller, Anthony Reina, Raghavan Jayakumar, and Malte Pietsch.
\newblock {COVID-QA}: A question answering dataset for {COVID}-19.
\newblock In {\em Proceedings of the 1st Workshop on {NLP} for {COVID-19} at
  {ACL} 2020}, Online, July 2020. Association for Computational Linguistics.

\bibitem{munirtowards}
Muhammad~Akhtar Munir, Muhammad~Haris Khan, M~Saquib Sarfraz, and Mohsen Ali.
\newblock Towards improving calibration in object detection under domain shift.
\newblock In {\em Advances in Neural Information Processing Systems}, 2022.

\bibitem{naeini2015obtaining}
Mahdi~Pakdaman Naeini, Gregory Cooper, and Milos Hauskrecht.
\newblock Obtaining well calibrated probabilities using {Bayesian} binning.
\newblock In {\em Proceedings of the AAAI Conference on Artificial
  Intelligence}, volume~29, 2015.

\bibitem{Nentidis_2022}
Anastasios Nentidis, Georgios Katsimpras, Eirini Vandorou, Anastasia Krithara,
  Antonio Miranda-Escalada, Luis Gasco, Martin Krallinger, and Georgios
  Paliouras.
\newblock Overview of~{BioASQ} 2022: The tenth {BioASQ} challenge
  on~large-scale biomedical semantic indexing and~question answering.
\newblock In {\em Lecture Notes in Computer Science}, pages 337--361. Springer
  International Publishing, 2022.

\bibitem{niculescu2005predicting}
Alexandru Niculescu-Mizil and Rich Caruana.
\newblock Predicting good probabilities with supervised learning.
\newblock In {\em Proceedings of the 22nd International Conference on Machine
  learning}, pages 625--632, 2005.

\bibitem{nixon2019measuring}
Jeremy Nixon, Michael~W Dusenberry, Linchuan Zhang, Ghassen Jerfel, and Dustin
  Tran.
\newblock Measuring calibration in deep learning.
\newblock In {\em CVPR Workshops}, volume~2, 2019.

\bibitem{pappas-etal-2020-biomrc}
Dimitris Pappas, Petros Stavropoulos, Ion Androutsopoulos, and Ryan McDonald.
\newblock {B}io{MRC}: A dataset for biomedical machine reading comprehension.
\newblock In {\em Proceedings of the 19th SIGBioMed Workshop on Biomedical
  Language Processing}, pages 140--149, Online, July 2020. Association for
  Computational Linguistics.

\bibitem{https://doi.org/10.48550/arxiv.2004.10796}
Jae~Sung Park, Chandra Bhagavatula, Roozbeh Mottaghi, Ali Farhadi, and Yejin
  Choi.
\newblock Visualcomet: Reasoning about the dynamic context of a still image,
  2020.

\bibitem{pasupat2015compositional}
Panupong Pasupat and Percy Liang.
\newblock Compositional semantic parsing on semi-structured tables.
\newblock {\em arXiv preprint arXiv:1508.00305}, 2015.

\bibitem{paz-argaman-tsarfaty-2019-run}
Tzuf Paz-Argaman and Reut Tsarfaty.
\newblock {RUN} through the streets: A new dataset and baseline models for
  realistic urban navigation.
\newblock In {\em Proceedings of the 2019 Conference on Empirical Methods in
  Natural Language Processing and the 9th International Joint Conference on
  Natural Language Processing (EMNLP-IJCNLP)}, pages 6449--6455, Hong Kong,
  China, Nov. 2019. Association for Computational Linguistics.

\bibitem{https://doi.org/10.48550/arxiv.2206.04045}
Michał Pietruszka, Michał Turski, Łukasz Borchmann, Tomasz Dwojak, Gabriela
  Pałka, Karolina Szyndler, Dawid Jurkiewicz, and Łukasz Garncarek.
\newblock Stable: Table generation framework for encoder-decoder models, 2022.

\bibitem{Powalski2021GoingFB}
Rafal Powalski, Łukasz Borchmann, Dawid Jurkiewicz, Tomasz Dwojak, Michal
  Pietruszka, and Gabriela Pałka.
\newblock Going full-tilt boogie on document understanding with
  text-image-layout transformer.
\newblock In {\em ICDAR}, 2021.

\bibitem{qi-etal-2022-dureadervis}
Le Qi, Shangwen Lv, Hongyu Li, Jing Liu, Yu Zhang, Qiaoqiao She, Hua Wu,
  Haifeng Wang, and Ting Liu.
\newblock $\textrm{DuReader}_{\textrm{vis}}$: A {C}hinese dataset for
  open-domain document visual question answering.
\newblock In {\em Findings of the Association for Computational Linguistics:
  ACL 2022}, pages 1338--1351, Dublin, Ireland, May 2022. Association for
  Computational Linguistics.

\bibitem{qin2022t5score}
Yiwei Qin, Weizhe Yuan, Graham Neubig, and Pengfei Liu.
\newblock T5score: Discriminative fine-tuning of generative evaluation metrics.
\newblock {\em arXiv preprint arXiv:2212.05726}, 2022.

\bibitem{raffel2020exploring}
Colin Raffel, Noam Shazeer, Adam Roberts, Katherine Lee, Sharan Narang, Michael
  Matena, Yanqi Zhou, Wei Li, Peter~J Liu, et~al.
\newblock Exploring the limits of transfer learning with a unified text-to-text
  transformer.
\newblock {\em J. Mach. Learn. Res.}, 21(140):1--67, 2020.

\bibitem{raghavan-etal-2021-emrkbqa}
Preethi Raghavan, Jennifer~J Liang, Diwakar Mahajan, Rachita Chandra, and Peter
  Szolovits.
\newblock emr{KBQA}: A clinical knowledge-base question answering dataset.
\newblock In {\em Proceedings of the 20th Workshop on Biomedical Language
  Processing}, pages 64--73, Online, June 2021. Association for Computational
  Linguistics.

\bibitem{rajpurkar2018know}
Pranav Rajpurkar, Robin Jia, and Percy Liang.
\newblock Know what you don't know: Unanswerable questions for squad.
\newblock {\em arXiv preprint arXiv:1806.03822}, 2018.

\bibitem{rajpurkar2016squad}
Pranav Rajpurkar, Jian Zhang, Konstantin Lopyrev, and Percy Liang.
\newblock Squad: 100,000+ questions for machine comprehension of text.
\newblock {\em arXiv preprint arXiv:1606.05250}, 2016.

\bibitem{roelofs2022mitigating}
Rebecca Roelofs, Nicholas Cain, Jonathon Shlens, and Michael~C Mozer.
\newblock Mitigating bias in calibration error estimation.
\newblock In {\em International Conference on Artificial Intelligence and
  Statistics}, pages 4036--4054. PMLR, 2022.

\bibitem{saxena-etal-2021-question}
Apoorv Saxena, Soumen Chakrabarti, and Partha Talukdar.
\newblock Question answering over temporal knowledge graphs.
\newblock In {\em Proceedings of the 59th Annual Meeting of the Association for
  Computational Linguistics and the 11th International Joint Conference on
  Natural Language Processing (Volume 1: Long Papers)}, pages 6663--6676,
  Online, Aug. 2021. Association for Computational Linguistics.

\bibitem{SIMPSON_1949}
E.~H. SIMPSON.
\newblock Measurement of diversity.
\newblock {\em Nature}, 163(4148):688--688, apr 1949.

\bibitem{slossberg2020calibration}
Ron Slossberg, Oron Anschel, Amir Markovitz, Ron Litman, Aviad Aberdam, Shahar
  Tsiper, Shai Mazor, Jon Wu, and R Manmatha.
\newblock On calibration of scene-text recognition models.
\newblock {\em arXiv preprint arXiv:2012.12643}, 2020.

\bibitem{smock2022pubtables}
Brandon Smock, Rohith Pesala, and Robin Abraham.
\newblock Pubtables-1m: Towards comprehensive table extraction from
  unstructured documents.
\newblock In {\em Proceedings of the IEEE/CVF Conference on Computer Vision and
  Pattern Recognition}, pages 4634--4642, 2022.

\bibitem{10.1145/3340531.3412760}
Tarc\'{\i}sio Souza~Costa, Simon Gottschalk, and Elena Demidova.
\newblock Event-qa: A dataset for event-centric question answering over
  knowledge graphs.
\newblock In {\em Proceedings of the 29th ACM International Conference on
  Information \& Knowledge Management}, CIKM '20, page 3157–3164, New York,
  NY, USA, 2020. Association for Computing Machinery.

\bibitem{kleisterStanislawekGWLK21}
Tomasz Stanislawek, Filip Gralinski, Anna Wr{\'{o}}blewska, Dawid Lipinski,
  Agnieszka Kaliska, Paulina Rosalska, Bartosz Topolski, and Przemyslaw Biecek.
\newblock Kleister: Key information extraction datasets involving long
  documents with complex layouts.
\newblock In {\em ICDAR}, volume 12821 of {\em Lecture Notes in Computer
  Science}, pages 564--579. Springer, 2021.

\bibitem{SlideVQA}
Ryota Tanaka, Kyosuke Nishida, Kosuke Nishida, Taku Hasegawa, Itsumi Saito, and
  Kuniko Saito.
\newblock Slidevqa: A dataset for document visual question answering on
  multiple images, 2023.

\bibitem{VisualMRC2021}
Ryota Tanaka, Kyosuke Nishida, and Sen Yoshida.
\newblock Visualmrc: Machine reading comprehension on document images.
\newblock In {\em AAAI}, 2021.

\bibitem{thorne-etal-2018-fever}
James Thorne, Andreas Vlachos, Christos Christodoulopoulos, and Arpit Mittal.
\newblock {FEVER}: a large-scale dataset for fact extraction and
  {VER}ification.
\newblock In {\em Proceedings of the 2018 Conference of the North {A}merican
  Chapter of the Association for Computational Linguistics: Human Language
  Technologies, Volume 1 (Long Papers)}, pages 809--819, New Orleans,
  Louisiana, June 2018. Association for Computational Linguistics.

\bibitem{tito2021document}
Rub{\`e}n Tito, Dimosthenis Karatzas, and Ernest Valveny.
\newblock Document collection visual question answering.
\newblock In {\em Document Analysis and Recognition--ICDAR 2021: 16th
  International Conference, Lausanne, Switzerland, September 5--10, 2021,
  Proceedings, Part II 16}, pages 778--792. Springer, 2021.

\bibitem{tito2022hierarchical}
Rub{\`e}n Tito, Dimosthenis Karatzas, and Ernest Valveny.
\newblock Hierarchical multimodal transformers for multi-page docvqa.
\newblock {\em arXiv preprint arXiv:2212.05935}, 2022.

\bibitem{tito2021icdar}
Rub{\`e}n Tito, Minesh Mathew, CV Jawahar, Ernest Valveny, and Dimosthenis
  Karatzas.
\newblock Icdar 2021 competition on document visual question answering.
\newblock In {\em International Conference on Document Analysis and
  Recognition}, pages 635--649. Springer, 2021.

\bibitem{trischler2016newsqa}
Adam Trischler, Tong Wang, Xingdi Yuan, Justin Harris, Alessandro Sordoni,
  Philip Bachman, and Kaheer Suleman.
\newblock Newsqa: A machine comprehension dataset.
\newblock {\em arXiv preprint arXiv:1611.09830}, 2016.

\bibitem{trivedi2017lc}
Priyansh Trivedi, Gaurav Maheshwari, Mohnish Dubey, and Jens Lehmann.
\newblock Lc-quad: A corpus for complex question answering over knowledge
  graphs.
\newblock In {\em International Semantic Web Conference}, pages 210--218.
  Springer, 2017.

\bibitem{vaicenavicius2019evaluating}
Juozas Vaicenavicius, David Widmann, Carl Andersson, Fredrik Lindsten, Jacob
  Roll, and Thomas Sch{\"o}n.
\newblock Evaluating model calibration in classification.
\newblock In {\em The 22nd International Conference on Artificial Intelligence
  and Statistics}, pages 3459--3467. PMLR, 2019.

\bibitem{https://doi.org/10.48550/arxiv.2204.07408}
Linyi Yang, Zhen Wang, Yuxiang Wu, Jie Yang, and Yue Zhang.
\newblock Towards fine-grained causal reasoning and qa, 2022.

\bibitem{yang2022multi}
Yuzhe Yang, Hao Wang, and Dina Katabi.
\newblock On multi-domain long-tailed recognition, generalization and beyond.
\newblock {\em arXiv preprint arXiv:2203.09513}, 2022.

\bibitem{yang-etal-2015-wikiqa}
Yi Yang, Wen-tau Yih, and Christopher Meek.
\newblock {W}iki{QA}: A challenge dataset for open-domain question answering.
\newblock In {\em Proceedings of the 2015 Conference on Empirical Methods in
  Natural Language Processing}, pages 2013--2018, Lisbon, Portugal, Sept. 2015.
  Association for Computational Linguistics.

\bibitem{yoshikawa-etal-2017-stair}
Yuya Yoshikawa, Yutaro Shigeto, and Akikazu Takeuchi.
\newblock {STAIR} captions: Constructing a large-scale {J}apanese image caption
  dataset.
\newblock In {\em Proceedings of the 55th Annual Meeting of the Association for
  Computational Linguistics (Volume 2: Short Papers)}, pages 417--421,
  Vancouver, Canada, July 2017. Association for Computational Linguistics.

\bibitem{you-etal-2022-end}
Chenyu You, Nuo Chen, Fenglin Liu, Shen Ge, Xian Wu, and Yuexian Zou.
\newblock End-to-end spoken conversational question answering: Task, dataset
  and model.
\newblock In {\em Findings of the Association for Computational Linguistics:
  NAACL 2022}, pages 1219--1232, Seattle, United States, July 2022. Association
  for Computational Linguistics.

\bibitem{yu2020reclor}
Weihao Yu, Zihang Jiang, Yanfei Dong, and Jiashi Feng.
\newblock Reclor: A reading comprehension dataset requiring logical reasoning.
\newblock In {\em International Conference on Learning Representations (ICLR)},
  April 2020.

\bibitem{zadrozny2002transforming}
Bianca Zadrozny and Charles Elkan.
\newblock Transforming classifier scores into accurate multiclass probability
  estimates.
\newblock In {\em Proceedings of the Eighth ACM SIGKDD International Conference
  on Knowledge Discovery and Data Mining}, pages 694--699, 2002.

\bibitem{zaheer2020big}
Manzil Zaheer, Guru Guruganesh, Kumar~Avinava Dubey, Joshua Ainslie, Chris
  Alberti, Santiago Ontanon, Philip Pham, Anirudh Ravula, Qifan Wang, Li Yang,
  et~al.
\newblock Big bird: Transformers for longer sequences.
\newblock {\em Advances in Neural Information Processing Systems},
  33:17283--17297, 2020.

\bibitem{zarharan-etal-2021-parsfever}
Majid Zarharan, Mahsa Ghaderan, Amin Pourdabiri, Zahra Sayedi, Behrouz
  Minaei-Bidgoli, Sauleh Eetemadi, and Mohammad~Taher Pilehvar.
\newblock {P}ars{FEVER}: a dataset for {F}arsi fact extraction and
  verification.
\newblock In {\em Proceedings of *SEM 2021: The Tenth Joint Conference on
  Lexical and Computational Semantics}, pages 99--104, Online, Aug. 2021.
  Association for Computational Linguistics.

\bibitem{zhang-etal-2021-noahqa-numerical}
Qiyuan Zhang, Lei Wang, Sicheng Yu, Shuohang Wang, Yang Wang, Jing Jiang, and
  Ee-Peng Lim.
\newblock {NOAHQA}: Numerical reasoning with interpretable graph question
  answering dataset.
\newblock In {\em Findings of the Association for Computational Linguistics:
  EMNLP 2021}, pages 4147--4161, Punta Cana, Dominican Republic, Nov. 2021.
  Association for Computational Linguistics.

\bibitem{zhang2021knowing}
Shujian Zhang, Chengyue Gong, and Eunsol Choi.
\newblock Knowing more about questions can help: Improving calibration in
  question answering.
\newblock {\em arXiv preprint arXiv:2106.01494}, 2021.

\bibitem{zhang2022nail}
Xinbo Zhang, Changzhi Sun, Yue Zhang, Lei Li, and Hao Zhou.
\newblock {NAIL}: A challenging benchmark for na{\textbackslash}''ive logical
  reasoning, 2022.

\bibitem{zheng2021global}
Xinyi Zheng, Douglas Burdick, Lucian Popa, Xu Zhong, and Nancy Xin~Ru Wang.
\newblock Global table extractor (gte): A framework for joint table
  identification and cell structure recognition using visual context.
\newblock In {\em Proceedings of the IEEE/CVF winter conference on applications
  of computer vision}, pages 697--706, 2021.

\bibitem{zhong2020image}
Xu Zhong, Elaheh ShafieiBavani, and Antonio Jimeno~Yepes.
\newblock Image-based table recognition: data, model, and evaluation.
\newblock In {\em Computer Vision--ECCV 2020: 16th European Conference,
  Glasgow, UK, August 23--28, 2020, Proceedings, Part XXI 16}, pages 564--580.
  Springer, 2020.

\bibitem{zhong2019publaynet}
Xu Zhong, Jianbin Tang, and Antonio~Jimeno Yepes.
\newblock Publaynet: largest dataset ever for document layout analysis.
\newblock In {\em 2019 International Conference on Document Analysis and
  Recognition (ICDAR)}, pages 1015--1022. IEEE, 2019.

\bibitem{PAMI_places}
Bolei Zhou, Agata Lapedriza, Aditya Khosla, Aude Oliva, and Antonio Torralba.
\newblock Places: A 10 million image database for scene recognition.
\newblock {\em {IEEE} Transactions on Pattern Analysis and Machine
  Intelligence}, 40(6):1452--1464, June 2018.

\bibitem{zhu2022towards}
Fengbin Zhu, Wenqiang Lei, Fuli Feng, Chao Wang, Haozhou Zhang, and Tat-Seng
  Chua.
\newblock Towards complex document understanding by discrete reasoning.
\newblock In {\em Proceedings of the 30th ACM International Conference on
  Multimedia}, pages 4857--4866, 2022.

\end{thebibliography}
}

\clearpage
\setcounter{page}{0}

\appendix
\onecolumn

\renewcommand{\contentsname}{\huge{\textcolor{red}{Supplementary Materials}}}

\etocdepthtag.toc{mtappendix}
\etocsettagdepth{mtchapter}{none}
\etocsettagdepth{mtappendix}{subsubsection}
\tableofcontents
\clearpage 

\twocolumn 

\section{Detailed Results Analysis}\label{app:results}

\subsection{Within Model Class Analysis}\label{app:analysis}

\subsubsection{Encoder vs. Decoder}


A key difference between encoder-only and (encoder-) decoder-based models is the ability to generate answers beyond the explicit document textual content. This is clearly reflected in the results for BigBird, Longformer, BERT, and LayoutLMv3, which score 
 $<10$ \ANLS{}\% on abstractive questions, whereas they have just average scores for extractive questions.
On \project{}, we can claim that a generative model is necessary given all considered question types.

Quite remarkably, while the human baseline demonstrates that humans find abstractive questions (\ANLS{} $\pm$82\%) easier than extractive questions (\ANLS{} $\pm$68\%), the reverse is true for all machine baselines. A potential confounder for these results could be the difference in output formatting for extractive vs. abstractive answers, which is hard to take into account with \ANLS{} evaluation.

\subsubsection{Incorporating Layout \& Vision}


When comparing T5 with and without 2D position embeddings on the diagnostic categories, we find the highest improvements on `evidence table or list',  `complexity simple', and `evidence plain'.

Our study with the proposed baselines shows that questions requiring visual evidence to be answered are an important future challenge for the vision community. To get further insights into models' performance on these questions, we calculate a weighted average of \ANLS{} over visual categories.
This reveals that GPT3 (4-shot) and T5-2d-large-8K obtain a tied score of (\ANLS{}=37\%), even though they only have access to the text. The human performance, on the other hand, is close to double (\ANLS{}=72\%), thus showing the need for better integration of the visual modality in DU models.


\subsubsection{Toward Long Document Processing}

\project{} clearly requires methods that can process long sequences, as evidenced by its average document length of $1832\pm2545$ tokens. This is particularly evident when comparing standard NLP QA methods like BERT-concat, which underperforms Longformer~\cite{beltagy2020longformer} and BigBird~\cite{zaheer2020big}, despite being the \textit{large} version. Experiments with T5 and T5-2D further support this claim, as extending the sequence length from 512 to 8192 leads to a $\sim5\%$ \ANLS{} improvement.

The exception is HiVT5~\cite{tito2022hierarchical}, which performs worse than the rest of the methods. This is due to the authors of HiVT5 performing a pre-training task of text denoising that helped to better model the [PAGE] tokens. This resulted in a better, compressed representation of the relevant information within a document conditioned by a question. Moreover, the authors also did extensive experimentation and found that 10 [PAGE] tokens per page were the best fit for the MP-DocVQA~\cite{tito2022hierarchical} dataset. We used similar hyperparameters, but \project{} might require better fine-tuning of [PAGE] tokens since the images are more visually rich with colored graphics and layouts.
The hierarchical processing of documents with a meaningful visual component is a promising avenue for future research.

\subsubsection{Diagnosis of LLM Results}\label{app:GPT}


The reasoning for including these LLMs as baselines stems from our question: ``Does advanced text understanding suffice for solving \project{}?".
Our results for diagnostic categories reveal some strengths and weaknesses of LLMs in the DocVQA task setting.

\textbf{Strengths}
GPT3 trumps all other tested models for list-type questions (\ANLS=36-40\%), which can be explained by the extractive nature of these questions.
After 4-shot fine-tuning, ChatGPT (4-shot) is better than all other tested baselines in answering not-answerable questions (\ANLS=77.45\%). This can partly explain the appeal of this particular GPT checkpoint in recent times.
GPT3 (4-shot) outperforms (\ANLS=52.51\%) other tested baselines on questions from the `complexity multi-hop' category such as \textsl{What city name appears the most often in the timetables?}.

\textbf{Weaknesses}
Compared to another (more simple text-only generative baseline, T5-base-512 (\ANLS=47\%), LLMs perform two times worse on abstractive questions (\ANLS=22\%). Closer analysis reveals that LLMs (even after 4-shot fine-tuning) predict abstractive questions to be \textit{Not-answerable} in $55\%$ of cases (in reality: 10\%). Operations such as arithmetic, counting, and comparisons remain generally elusive skills (<25\%\ANLS). 

Both LLMs we tested scored significantly lower than the human baseline in questions that require visual understanding, with an average \ANLS~score of 21\%. This is understandable because these are text-only models.

While LLMs' zero-shot performance is relatively high, we note that \project{} consists of public-license documents from the web, which potentially might have been included in the LLMs' pre-training corpus. 

\subsection{Assessing Confidence}\label{app:conf}

\ECE{} measures calibration of confidence, whereas \AURC{} assesses both performance and confidence ranking \cite{jaeger2023a} (more detail \Cref{app:metrics}). The latter results in an appropriate metric to select the best model in real-world applications, where wrong predictions can yield undesired scenarios, which could be prevented by manually revising low-confidence answers.

Interestingly, T5-base-512 scores better on calibration (\ECE=10.82) than T5-2D-large-8K, the baseline with the highest \ANLS, yet worse calibration (\ECE=14.4). In general, it seems calibration worsens when extending the maximum sequence length, whereas adding 2D position embeddings only positively affects 
 \ANLS. From the baselines tested, T5-2D-large-8K achieves the highest \AURC. 

Another interesting result comes from analyzing the calibration of models evaluated using the \textit{Concat} strategy vs. \textit{Max Conf.} strategy. In the main paper, we reported results for the model with the relative best \ANLS{}. Thanks to our varied set of evaluation metrics, we discover that \textit{Max Conf.} overall results in poor calibration (see \Cref{tab:logits-vs-concat}), whereas considering \ANLS, there is not always a clear winning strategy.
This shows that predicting each page separately and necessarily assuming conditional independence across pages is not a reliable strategy for multipage DocVQA. 

\begin{table}
    \centering
    \resizebox{\columnwidth}{!}{%
    \begin{tabular}{c|ccc}
\toprule
Model  &  \ANLS & \ECE & \AURC  \\ 
\midrule
BertQA MPDocVQA Concat  &  29.8 & \textbf{13.83} & \textbf{43.28}  \\ 
BertQA MPDocVQA MaxConf  &  \textbf{32.18} & 28.93 & 48.73  \\ \midrule
BigBird MPDocVQA Concat  &  \textbf{30.67} & \textbf{25.07} & \textbf{47.2}  \\ 
BigBird MPDocVQA MaxConf  &  29.38 & 50.79 & 56.81  \\ \midrule
LayoutLMv3 MPDocVQA Concat  &  22.61 & \textbf{13.19} & 57.11  \\ 
LayoutLMv3 MPDocVQA MaxConf  &  \textbf{25.27} & 31.31 & 58.54  \\ \midrule
Longformer MPDocVQA Concat  &  \textbf{33.45} & \textbf{22.21} & 45.83  \\ 
Longformer MPDocVQA MaxConf  &  28.67 & 48.6 & 58.11  \\ \midrule
T5 MPDocVQA Concat  &  34.37 & \textbf{18.97} & 47.31  \\ 
T5 MPDocVQA MaxConf  &  \textbf{37.56} & 23.73 & \textbf{46.69}  \\ \midrule
T5-base Concat-0  &  \textbf{25.62} & \textbf{20.05} & 62.25  \\ 
T5-base MaxConf-0  &  22.21 & 39.47 & \textbf{58.89}  \\
\bottomrule
    \end{tabular}}
    \caption{
    Comparison of baselines using Concat or Max Conf strategies.}
    \label{tab:logits-vs-concat}
\end{table}

\section{Baseline Experiments Setup}\label{app:experiments}


In this Section, we describe the implementation details for the architectures and inference methods used in our benchmark.

\subsection{Hyperparameter Defaults}

\begin{table}[]
\small
\centering
\begin{tabular}{lccc} 
\toprule
Hyper-Parameter & T5 & T5+2D & HiVT5 \\ \midrule
Epochs & 10 & 10 & 10 \\
\begin{tabular}[c]{@{}l@{}}Warm-up\\ (iterations)\end{tabular} & 1000 & 250 & 1000 \\
Optimizer & Adam, AdamW & Adafactor & Adam \\
Gradient acc. & False & 8 & False \\
Lower case & True & True & True \\
Max. Seq. Length & 512, 8192 & 512, 8192 & 20480 \\
\begin{tabular}[c]{@{}l@{}}Generation\\ (Max. Tokens)\end{tabular} & 100 & 100 & 50 \\
Batch size & 3 & 8 & 1 \\
Learning rate & 1E-04, 2E-04 & 2E-04 & 2E-04 \\
\begin{tabular}[c]{@{}l@{}}Training time\\ (per epoch)\end{tabular} & 1h, 10h & 1.5h, 5h & 10h \\
GPU Hardware & \begin{tabular}[c]{@{}c@{}}TITAN RTX,\\ A100\end{tabular} & \begin{tabular}[c]{@{}c@{}}A100\\ (80GB)\end{tabular} & \begin{tabular}[c]{@{}c@{}}TITAN RTX\\ (24GB)\end{tabular} \\
\bottomrule
\end{tabular}
\caption{Hyperparameters used for fine-tuning \texttt{T5}, \texttt{T5-2D} and \texttt{HiVT5} on \project{}. When two values are placed in a single column, they refer to the model's versions with 512 and 8192 input sequence length, respectively.}
\label{tab:finetune_params}
\end{table}

\noindent Refer to Table~\ref{tab:finetune_params}.

\subsection{Generative LLM Prompt Fine-tuning}

The performance of GPT3.5 models was assessed in two settings: 0-shot and 4-shot.  In the 0-shot setting, the prompt included instructions similar to those provided to annotators to teach them how to annotate. In the 4-shot setting, the prompt was enhanced with the content of a single document from the training set along with four questions of different types (extractive, abstractive, list, and not answerable) to better gauge the models' abilities.

\begin{tcolorbox}[colback=lightgray,colframe=black,width=\columnwidth,arc=5mm,title=\large Few-shot Prompts]
\footnotesize{
\textbf{Document:}

<content of single training set document>

--

\textbf{Question:} <extractive question>

\textbf{Answer:} <extractive answer>

--

\textbf{Question:} <abstractive question>

\textbf{Answer:} <abstractive answer>

--

\textbf{Question:} <list question>

\textbf{Answer:} <list item> | <list item>

--

\textbf{Question:} <not-answerable question>

\textbf{Answer:} none

--

\textbf{Document:}

<content of evaluated document>

--

Questions and answers pairs to above document:

Answers contains either:

- a span inside of document

- a list of spans inside of document (each span should be separated by "|")

- not exist explicitly as span of document (the answer should be freely generated text)

- question couldn't be answered (the answer should be "none")

\textbf{Question:} <question>

\textbf{Answer:} \_\_\_\_\_\_
}
\end{tcolorbox}

The 0-shot prompt is analogous to the 4-shot prompt, but the key distinction is that it lacks the first document and the example question-and-answer pairs. 

For the inference process, we utilized the prompt completion default settings outlined in the OpenAI documentation, with the exception of the temperature parameter, which was lowered to a value of 0.0. This adjustment was made to ensure that the output would be more deterministic and focused, with less emphasis on generating creative variations.

Only after our prompting experiments had been completed, we realized the opportunity to assess confidence estimation using chained prompts (\textsl{Please give a confidence between 0 and 1 about how certain you are this is the answer.}) as in \cite{projectid}. Since we did not save our dialogue states and considered the expenses, we leave this for future work. 

\subsection{Confidence Estimation}\label{app:confidence}

This Subsection details confidence scoring functions for the baselines, as this is not reported in standard practice.

We define \textit{confidence} as the predicted probability of the top-1 prediction, often arising as the largest value from softmax normalization of logits from a final model layer (head).

\textbf{Encode}r-based models will output logits for all possible start and end positions of the answer within the provided context. While the predicted answer of such a span prediction architecture will come from the highest valid (no negative span) combination of the sum of a start and end logit, the predicted answer confidence can be obtained by the following procedure ($BS$: batch size and $S$: sequence length): (to be added in next version)






\textbf{Decoder}-based models are not restricted to spans and can output an arbitrary, though often controllable, amount of text tokens, indicated as $S^{\prime}$. The logits at the final layer take the shape of $BS \times S^{\prime} \times V$, where $V$ is the tokenizer's vocabulary size (32.1K for T5-base). The following confidence estimation procedure 
is applied for decoder outputs: (to be added in next version)





\subsection{Evaluation metrics}\label{app:metrics}

All metric implementations (\ANLS{}, \ECE{}, \AURC{}) are made available as a standalone repository. 
Additionally, we provide an online service where researchers can evaluate their methods against a blind (questions-only) test dataset.
Below, we expound on the implementation details of the metrics and motivate design choices.

\subsubsection{\ANLS}

Average Normalized Levenshtein Similarity (\ANLS) is a metric introduced in \cite{biten2019scene}, which was then extended \cite{tito2021document} to support \textit{lists} and be invariant to the order of provided answers.
We adapt the underlying Levenshtein Distance metric \cite{levenshtein1966binary} to support \textit{not-answerable} questions,  
 $\operatorname{NA}(G) = \mathbb{I} [\operatorname{type}(G) = \text{ not-answerable }]$ (see \Cref{eq:ls}).

Consider for simplicity, the evaluation of a single non-list ground truth answer $G$ and prediction $\hat{P}$, each with string lengths $|G|$ and $|\hat{P}|$, respectively.
\begin{gather}
\operatorname{LD}(G, \hat{P})=\left\{\begin{array}{l}
1 \quad \text { if }\operatorname{NA}(G) \land |\hat{P}| > 0 , \\
0 \quad \text { if }\operatorname{NA}(G) \land |\hat{P}| = 0 , \\
|G| \quad  \text { if }|\hat{P}|=0 , \\
\operatorname{LD}(\operatorname{tail}(G), \operatorname{tail}(\hat{P})) \quad \text { if } G[0]=\hat{P}[0], \\
1+\min \left\{\begin{array}{ll}
\operatorname{LD}(\operatorname{tail}(G), \hat{P}) & \\
\operatorname{LD}(G, \operatorname{tail}(\hat{P})) \text {  , otherwise} \\
\operatorname{LD}(\operatorname{tail}(G), \operatorname{tail}(\hat{P})) &
\end{array}\right.
\end{array}\right.
\label{eq:ls}
\end{gather}

The normalized similarity metric is then defined as 
$$\operatorname{NLS}(G, \hat{P}) = \frac{1-\operatorname{LD}(G, \hat{P})}{\max(1, |G|, |\hat{P}|)}.$$

Given multiple ground truth answer variants $G = \{g_1, g_2, ...\}$ and a predicted answer for $\hat{P}_{Q_i}$ for each question $Q$ in the test set of size $N$, we define the complete metric as follows: \begin{gather}
\mathrm{ANLS}=\frac{1}{N} \sum_{i=0}^N\left(\max_{g \in G_i} s\left(g, \hat{P}_{Q_i}\right)\right) \\
s\left(g, \hat{P}_{Q_i}\right)=\left\{\begin{array}{ll}
\operatorname{NLS}\left(g, \hat{P}_{Q_i}\right) & \text { if } \operatorname{NLS}\left(g, \hat{P}_{Q_i}\right)\geqslant\tau \\
0 & \text { if } \operatorname{NLS}\left(g, \hat{P}_{Q_i}\right) < \tau
\end{array}\right.
,\end{gather} where we follow prior literature \cite{biten2019scene,tito2021document} in setting the threshold $\tau=0.5$. 

In the case of a \textit{list}-type question,  Hungarian matching is performed following \cite{tito2021document} according to $\mathrm{NLS}$ between each ground truth answer part and each prediction answer part.

While \ANLS{} can account for shortcomings of OCR and formatting issues, evaluation of generated text is notoriously complex \cite{qin2022t5score} and requires more research. 

\subsubsection{\ECE}

Expected Calibration Error (\ECE) is a default metric to evaluate top-1 prediction miscalibration. It measures the $\mathcal{L}_p$ norm difference between a model’s posterior and the true likelihood of being correct, as formally defined below: $$ ECE_p(f)^p= \mathbb{E}_{(X,Y)} \left[\|\mathbb{E}[Y = \hat{y} \mid f(X) = \hat{p}] - f(X)\|^p_p\right],$$ where $\hat{y} = \arg\max_{y'}[f(X)]_y'$ is a class prediction with associated posterior probability $\hat{p}= \max_{y'}[f(X)]_y'$.

In our setting, the exact accuracy condition $\mathbb{I}[Y = \hat{y}]$ is replaced by $\mathbb{I}[\mathrm{ANLS}(y, \hat{y})>\tau]$. Prior work \cite{munirtowards} already introduced the strategy of thresholding continuous quality scores (in the case of IOU larger than $\tau$) in order to be able to estimate \ECE. 

In practice, \ECE{} is implemented as a histogram binning estimator that discretizes predicted probabilities into ranges of possible values (bins) for which conditional expectation can be estimated.
In order to minimize the drawbacks inherited from histogram binning, as suggested by the literature \cite{nixon2019measuring,vaicenavicius2019evaluating,kumar2019calibration,roelofs2022mitigating}, we apply an equal-mass binning scheme with 100 bins (close to $\sqrt{N}$).

\subsubsection{\AURC}

Area-Under-Risk-Coverage-Curve (\AURC) \cite{geifman2017selective,jaeger2023a} measures the possible trade-offs between coverage (proportion of test set\%) and risk (error \% under given coverage). It assumes predictions to come with a confidence estimate. The curve can be obtained by sorting all confidence estimates and evaluating risk from high to low, together with their respective correctness (typically based on exact match).

Similar to \ECE{} as defined above, we apply \ANLS{} thresholding instead. Formulated this way, the best possible \AURC{} is constrained by the model's test error (1-\ANLS{}) and the number of test instances. 
We have extended the very detailed implementation of \cite{jaeger2023a}, to which we refer for further information.
On a final note, \AURC{} might be more sensible for evaluating highly-accurate settings (e.g., 90\% accuracy), where risk can be better controlled (as it is typically a business decision to decide tolerance to mistakes).


\section{Qualitative Examples}\label{app:samples}

As is customary, we provide some interesting, handpicked test set examples with predictions from some of the baselines in our study.

\noindent\begin{minipage}{\linewidth}
\paragraph{Low complexity.} \textsl{Where the document has been printed?}
\vspace{1pt}\\
Simple, extractive question, plain-text evidence.

\begin{figure}[H]
    \includegraphics[width=\linewidth]{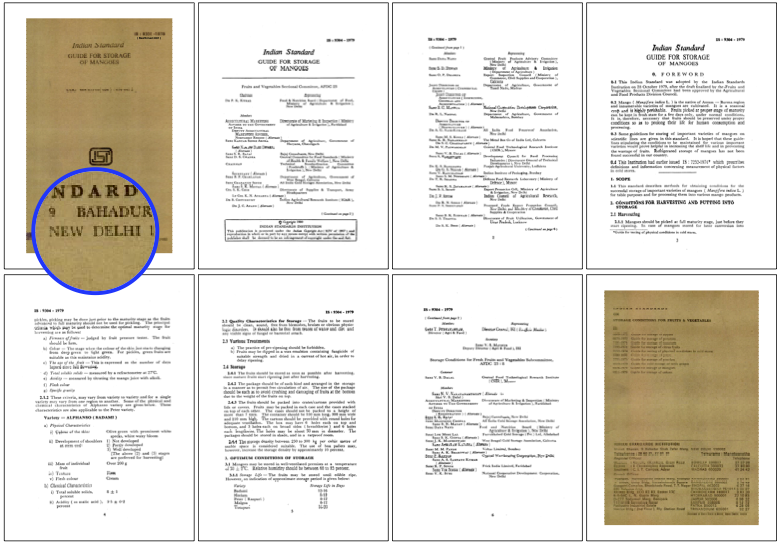}
    \vspace{-1mm}\\
    \begin{tabular}{lp{3cm}rr}
        \toprule
        Source & Answer & ANLS & Conf. \\
        \midrule
        Ground truth & \textit{New Delhi, India} \\
        Human & \textit{India} & $0.0$ & --- \\
        \midrule
        T5 & \textit{IS : 9304 - 1979} & $0.0$ & $0.56$ \\
        ChatGPT & \textit{The document does not mention where it has been printed.} & $0.0$ & --- \\
        GPT3 & \textit{Bela Pack n Print. New Delhi, India} & $0.0$ & --- \\
        T5-2D & \textit{New Delhi, India} & $1.0$ & $0.09$ \\
        HiVT5 & \textit{Page 1} & $0.0$ & $0.18$ \\
        Longformer & new delhi, india & $1.0$ & $0.72$  \\
        \bottomrule
    \end{tabular}
\end{figure}
\vspace{1pt}
\end{minipage}

\noindent\begin{minipage}{\linewidth}

\paragraph{High complexity.} \textsl{Is there any redacted section on the document?}
\vspace{2mm}
Abstractive question that requires knowledge about possible document elements.

\begin{figure}[H]
    \includegraphics[width=\linewidth]{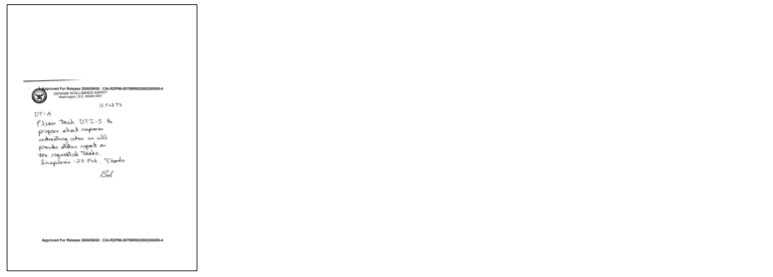}
    \vspace{-1mm}\\
    \begin{tabular}{lp{3cm}rr}
        \toprule
        Source & Answer & ANLS & Conf. \\
        \midrule
        Ground truth & \textit{No} \\
        Human & \textit{No} & $1.0$ & --- \\
        \midrule
        T5 & \textit{yes} & $0.0$ & $0.17$ \\
        ChatGPT & \textit{[Not-answerable]} & $0.0$ & ---\\
        GPT3 & \textit{[Not-answerable]} & $0.0$ & ---\\
        T5-2D & \textit{No} & $1.0$ & $0.43$ \\
        HiVT5 & \textit{Yes} & $0.0$ & $0.55$ \\
        LayoutLMv3 & \textit{approved for release} & $0.0$ & $0.01$ \\
        \bottomrule
    \end{tabular}
\end{figure}

\end{minipage}

\noindent\begin{minipage}{\linewidth}
\paragraph{Requires arithmetic.} \textsl{What is the difference between how much Operator II and Operator III makes per hour?}
\vspace{1pt}
The question requires table comprehension, determining relevant values, dividing extracted integers, and correcting the subject-verb agreement.

\begin{figure}[H]
    \includegraphics[width=0.7\linewidth]{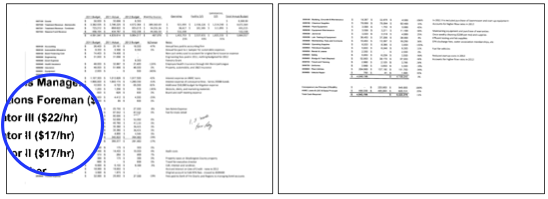}
    \vspace{-1mm}\\
    
    \begin{tabular}{lp{3.1cm}rr}
        \toprule
        Source & Answer & ANLS & Conf. \\
        \midrule
        Ground truth & \textit{\$5} \\
        Human & \textit{\$5} & $1.0$  & --- \\
        \midrule
        T5 & \textit{200} & $0.0$  & $0.28$\\
        ChatGPT & \textit{\$5 per hour.} & $0.0$ & --- \\
        GPT3 & \textit{Operator II (\$17/hr) | Operator III (\$22/hr)} & $0.0$ & --- \\
        T5-2D & \textit{[Not-answerable]} & $0.0$ & $0.31$ \\
        HiVT5 & \textit{[Not-answerable]} & $0.0$ & $0.15$ \\
        \bottomrule
    \end{tabular}
\end{figure}
\vspace{1mm}
\end{minipage}

\noindent\begin{minipage}{\linewidth}
\paragraph{Visual evidence (chart).} \textsl{What is the maximum percentage of the blue graph line on page 8?}
\vspace{2mm}
A highly demanding question that requires simultaneous competency of visual comprehension (locating chart and line color), navigating through layout (determining adequate page), and numerical comparison (deciding on the highest value).

\begin{figure}[H]
    \includegraphics[width=\linewidth]{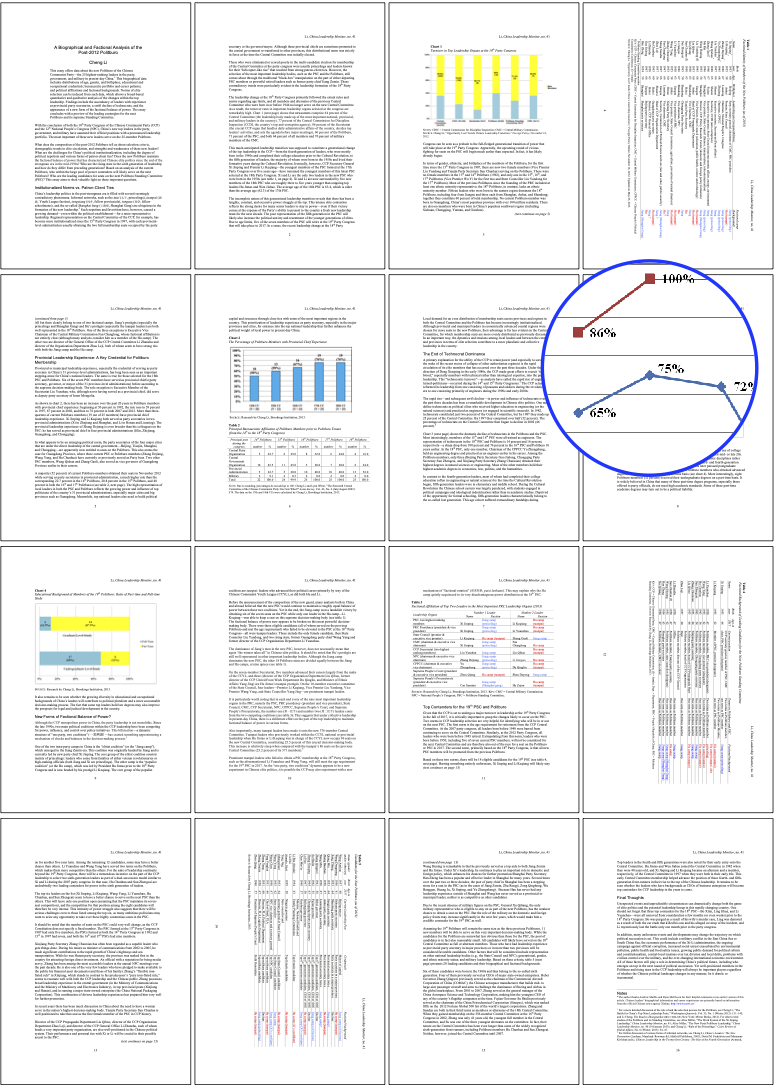}
    \vspace{-1mm}
    
    \begin{tabular}{llrr}
        \toprule
        Source & Answer & ANLS & Conf. \\
        \midrule
        Ground truth & \textit{75\%} \\
        Human & \textit{75} & $0.7$ & --- \\
        \midrule
        T5 & \textit{76} & $0.0$ & $0.25$ \\
        ChatGPT & \textit{[Not-answerable]} & $0.0$ & ---\\
        GPT3 & \textit{76\%} & $0.7$ & --- \\
        T5-2D & \textit{32.0} & $0.0$ & $0.00$ \\
        HiVT5 & \textit{45\%} & $0.7$ & $0.05$ \\
        BigBird & \textit{32 }& $0.0$ & $0.47$  \\
        LayoutLMv3 & \textit{80\%} & $0.0$ & $0.15$  \\
        \bottomrule
    \end{tabular}
\end{figure}
\end{minipage}

\noindent\begin{minipage}{\linewidth}
\paragraph{Visual evidence (handwriting).} \textsl{What is the handwritten date on page 1?}
\vspace{1mm}
The question requires visual comprehension (recognition of handwriting) and layout navigation (determining the adequate page).

\begin{figure}[H]
    \includegraphics[width=\linewidth]{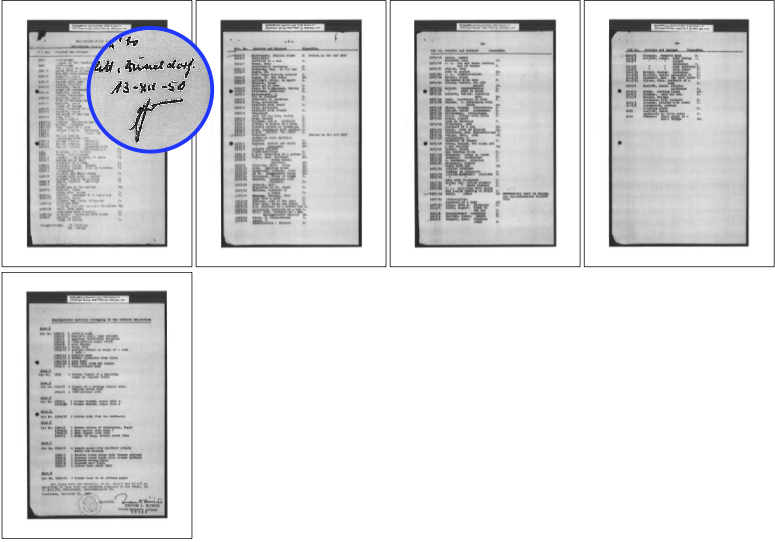}
    \vspace{-1mm}\\
    \begin{tabular}{llrr}
        \toprule
        Source & Answer & ANLS & Conf. \\
        \midrule
        Ground truth & \textit{13-XII-50} \\
        Human & \textit{13-XII-50} & $1.0$ & --- \\
        \midrule
        T5 & \textit{1977-01-01} & $0.0$ & $0.24$  \\
        ChatGPT & \textit{[Not-answerable]} & $0.0$ & ---\\
        GPT3 & \textit{15 December 1950} & $0.0$ & --- \\
        T5-2D & \textit{1950-12-15} & $0.0$ & $0.24$ \\
        HiVT5 & \textit{1977-07-01} & $0.0$ & $0.11$ \\
        BERTQA & \textit{2006 / 1} & $0.0$ & $0.5$  \\
        \bottomrule
    \end{tabular}
\end{figure}
\vspace{5mm}
\end{minipage}

\noindent\begin{minipage}{\linewidth}
\paragraph{Requires counting.} \textsl{How many pages have a signature?}

\vspace{1mm}
The question requires visual comprehension (recognition of signature), knowledge about  layout, and counting.

\begin{figure}[H]
    \includegraphics[width=\linewidth]{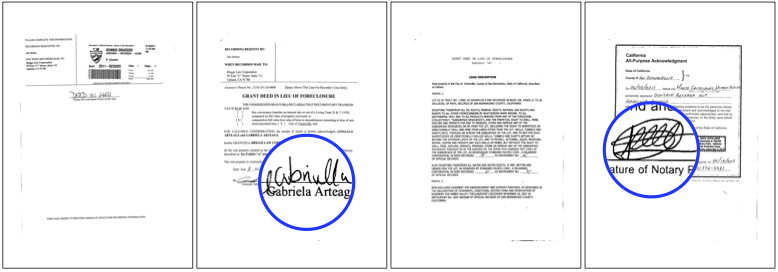}
    \vspace{-6mm}
    
    \begin{tabular}{llrr}
        \toprule
        Source & Answer & ANLS & Conf. \\
        \midrule
        Ground truth & \textit{2} \\
        Human & \textit{2} & $1.0$ & --- \\
        \midrule
        T5 & \textit{1} & $0.0$ & $0.01$  \\
        ChatGPT & \textit{4} & $0.0$ & --- \\
        GPT3 & \textit{[Not-answerable]} & $0.0$ & ---\\
        T5-2D & \textit{4} & $0.0$ & $0.69$ \\
        HiVT5 & \textit{4} & $0.0$ & $0.41$ \\
        \bottomrule
    \end{tabular}
\end{figure}
\end{minipage}

\noindent\begin{minipage}{\linewidth}

\paragraph{Visual evidence (map), multi-hop.} \textsl{Which states don't have any marijuana laws?}
\vspace{2mm}
The multi-hop question requires visually comprehending the map and linking knowledge from its legend with depicted regions.

\begin{figure}[H]
    \includegraphics[width=\linewidth]{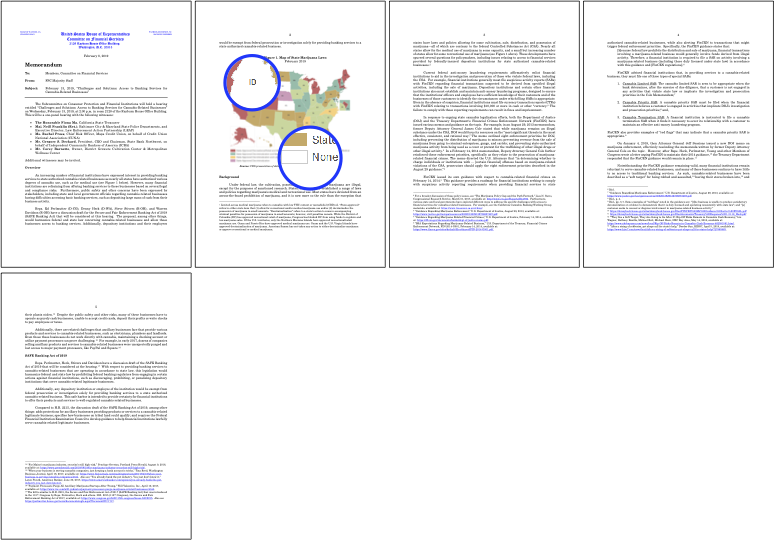}
    \vspace{-4mm}\\
    
    \begin{tabular}{lp{3.1cm}rr}
        \toprule
        Source & Answer & ANLS & Conf. \\
        \midrule
        Ground truth & \textit{ID | SD | KS} \\
        Human & \textit{ID | SD | KS} & $1.0$ & --- \\
        \midrule
        T5 & \textit{WA ME MT ND MN OR VT ID NH SD WI NY MA MI} & $0.0$ & $0.28$ \\
        ChatGPT & \textit{[Not-answerable]} & $0.0$ & ---- \\
        GPT3 & \textit{American Samoa} & $0.0$ & ---- \\
        T5-2D & \textit{i} & $0.0$ & $0.03$ \\
        HiVT5 & \textit{-} & $0.0$ & $0.02$ \\
        \bottomrule
    \end{tabular}
\end{figure}
\end{minipage}



\clearpage

\section{Additional Dataset Statistics}\label{app:statistics}

\subsection{Answer Types}\label{app:correlation_heatmap}
\begin{figure}[h]
    \centering
\includegraphics[width=0.65\linewidth]{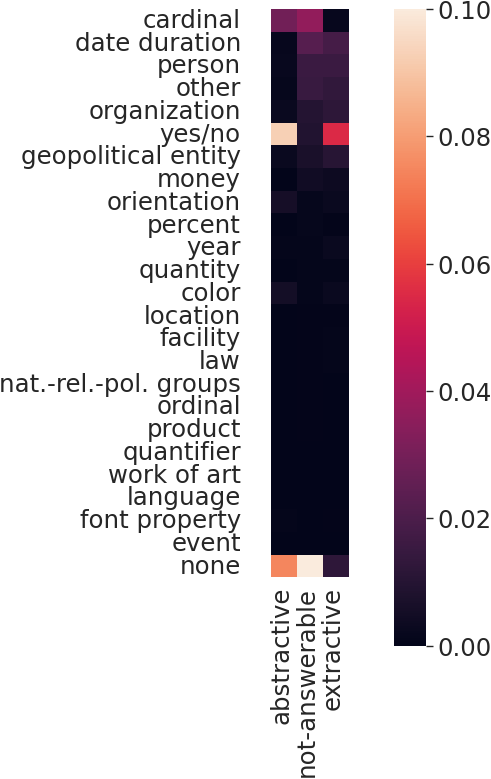}
    \caption{Answer types correlation heatmap. Results obtained with Cramer's V coefficient. Note that values on the scale are below 0.1, suggesting a lack of correlation.}
    \label{fig:atype_heatmap}
\end{figure}

\Cref{fig:atype_heatmap} shows that there are barely any correlations between question type and answer type, except for the most expected ones (e.g. `None' answers and `Not answerable' questions), by means of Cramer's V coefficient. For instance, date and duration types of answers are equally likely for both extractive and abstractive questions.

\begin{figure}[h]
    \centering
\includegraphics[width=1.1\linewidth]{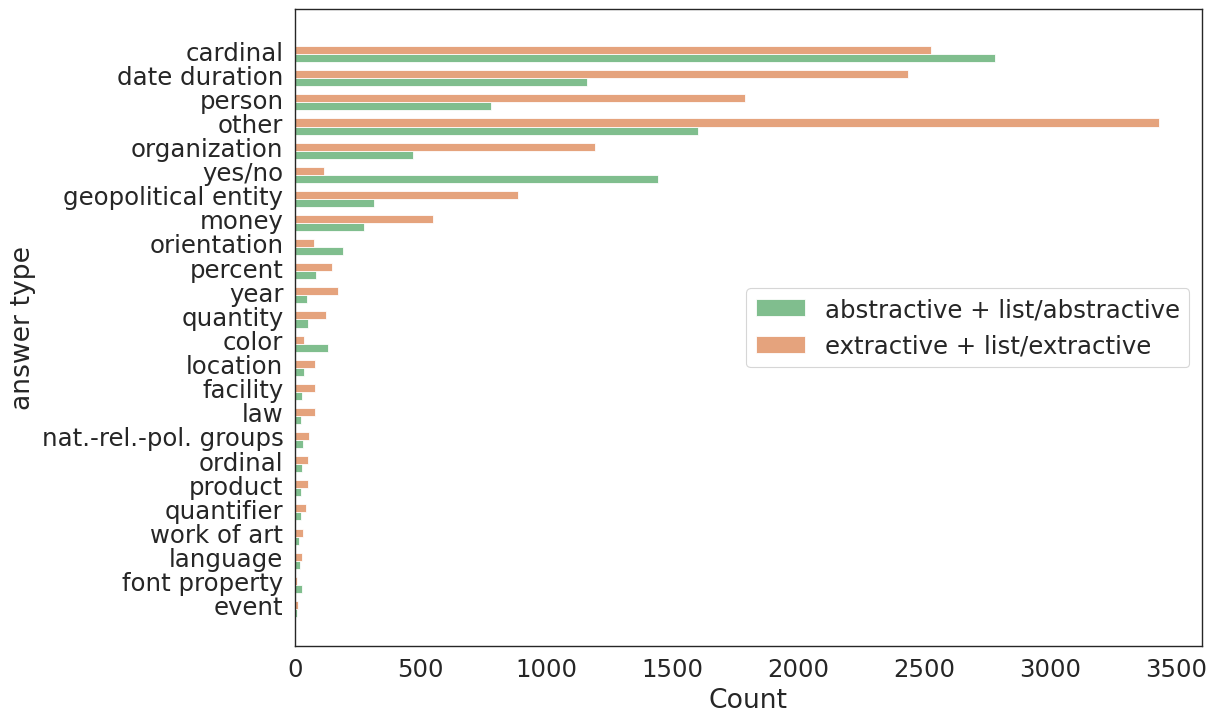}
    \caption{Answer type distribution per question type in \project.}
    \label{fig:atype_dist}
\end{figure}

\begin{figure}[h]
    \centering
\includegraphics[width=1.1\linewidth]{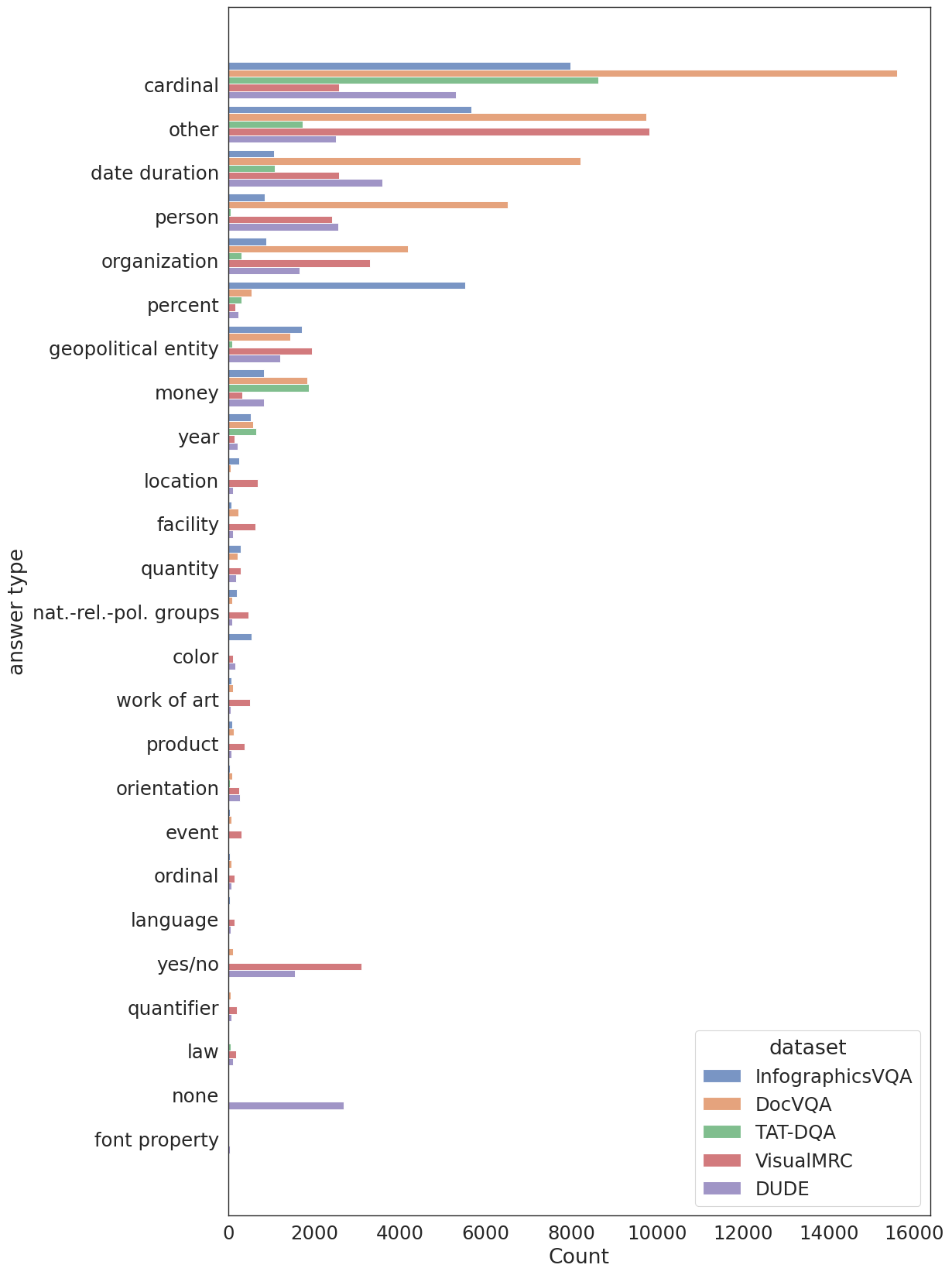}
    \caption{Answer type distribution per dataset, sorted in descending order of total answer type occurrences. We have found: 13 answer types in TAT-DQA; 20 answer types in InfographicsVQA and SP-DocVQA, 23 answer types in VisualMRC, and 24 answer types in \project{}}
    \label{fig:atype_dist_all}
\end{figure}

\Cref{fig:atype_dist} shows the answer type distribution per question type in \project{}, followed by a comparison to answer type distributions in related DocVQA datasets (\Cref{fig:atype_dist_all}). 

\subsection{Dataset Diversity}
\begin{figure*}[bth]
    \centering
    \includegraphics[width=0.8\linewidth,trim={2.5cm 1.5cm 0.8cm 1.6cm},clip]{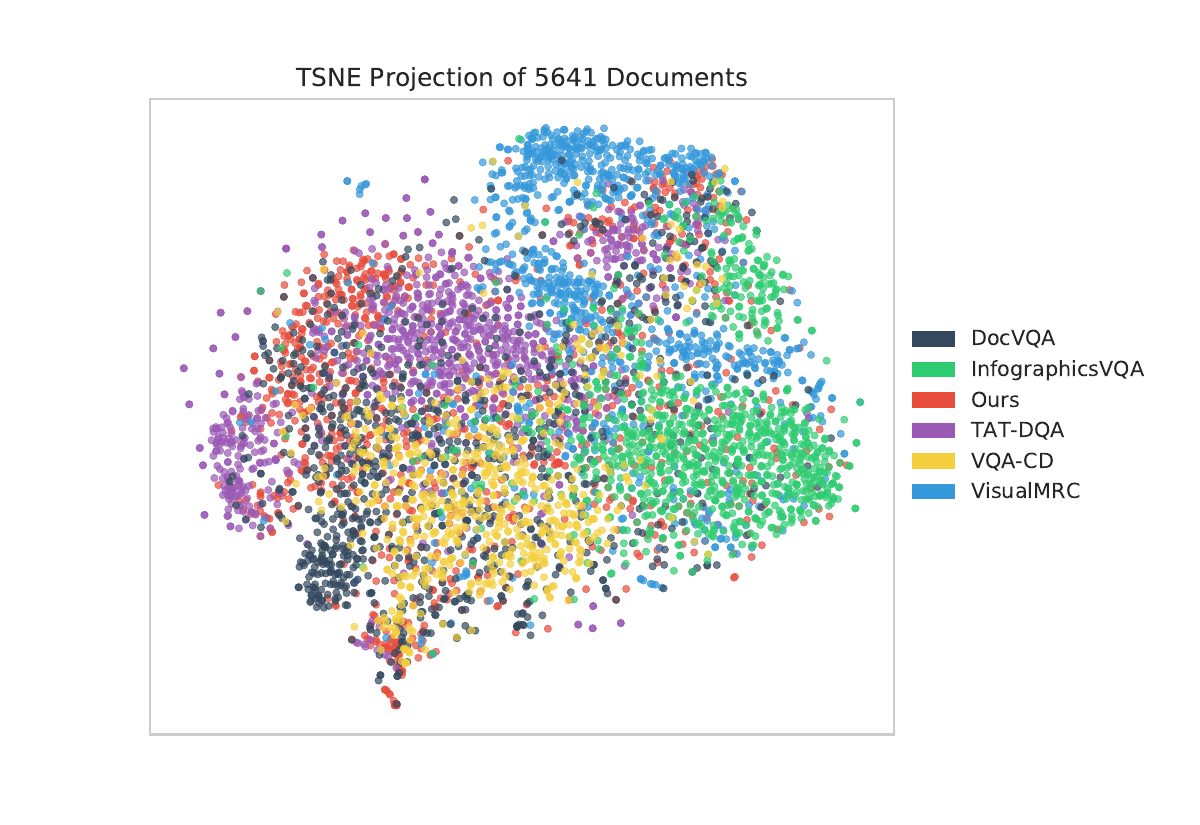}
    \caption{Visualization of document image similarities between samples from different datasets (t-SNE over ResNet101 features of 1k documents, first pages only).}
    \label{fig:layout}
\end{figure*}

Similar to the text-based comparison, \Cref{fig:layout} visualizes the diversity of the visual embeddings of all documents' first pages in \project, relative to those from other DocVQA datasets. 



\end{document}